\documentclass{IEEEtran}
\usepackage{subcaption}
\usepackage[utf8]{inputenc} 
\usepackage[T1]{fontenc}    
\usepackage{rotating}
\usepackage{lineno}
\usepackage[markup=default]{changes}
\usepackage{comment}
\usepackage{url}            
\usepackage{array}
\usepackage{ragged2e}
\usepackage{booktabs}       
\usepackage{longtable}
\usepackage{amsfonts}       
\usepackage{nicefrac}       
\usepackage{microtype}      
\usepackage{multirow}
\usepackage{colortbl}
\usepackage{times}
\usepackage{tabularray}
\UseTblrLibrary{booktabs}
\usepackage{latexsym}
\usepackage{makecell}
\usepackage{epsfig}
\usepackage{graphicx}
\usepackage{float} 
\usepackage{amsmath}
\usepackage{amssymb}
\usepackage{bbding}
\usepackage{color}
\usepackage{soul}
\usepackage{wrapfig}
\usepackage{transparent}
\usepackage{tabularx}
\usepackage{tabu}
\usepackage{algorithmic}
\usepackage[ruled,linesnumbered]{algorithm2e}
\usepackage[most]{tcolorbox}
\usepackage{xcolor} %
\newcommand{\rev}[1]{#1}
\usepackage{adjustbox}
\usepackage{fontawesome}
\usepackage{CJK}
\usepackage{caption}  
\usepackage{orcidlink}

\usepackage{hyperref}
\lstdefinestyle{cleanJson}{
    basicstyle=\ttfamily,
    breaklines=true,          %
    breakatwhitespace=true,   %
    breakindent=20pt,         %
    postbreak=\space,         %
}
\usepackage{inconsolata}
\usepackage{pifont}

\definecolor{softblue}{rgb}{0.88, 0.95, 1.0} %
\definecolor{softyellow}{rgb}{0.98, 0.98, 0.82} %
\definecolor{green}{rgb}{0, 1, 0}
\definecolor{REBUTTAL}{rgb}{1, 0, 0}

\newtcolorbox{prompt}[1]{
    enhanced,
    breakable=true,
    colback=gray!3,
    colframe=black!70,
    boxrule=0.5pt,
    arc=3mm,
    left=10pt,right=10pt,top=4pt,bottom=4pt,
    boxsep=5pt,
    fonttitle=\bfseries,
    title=#1,
}
\newtcolorbox{example}[1]{
    enhanced,
    breakable=true,
    colback=blue!3,
    colframe=black!70,
    boxrule=0.5pt,
    arc=3mm,
    left=10pt,
    right=10pt,
    boxsep=5pt,
    fonttitle=\bfseries,
    title=#1
}

\newtcolorbox{case}[1]{
    enhanced,
    breakable=true,
    colback=red!3,
    colframe=black!70,
    boxrule=0.5pt,
    arc=3mm,
    left=10pt,
    right=10pt,
    boxsep=5pt,
    fonttitle=\bfseries,
    title=#1
}

\makeatletter
\newcommand{\namelabel}[1]{%
  \phantomsection
  \renewcommand{\@currentlabel}{#1}
  \label{#1}
}
\makeatother

\usepackage{eso-pic}
\newcommand{\arxivnotice}{%
  \AddToShipoutPictureFG*{%
    \AtPageUpperLeft{%
      \put(\LenToUnit{0.5\paperwidth},\LenToUnit{-1.2cm}){%
        \makebox[0pt][c]{\parbox{0.85\paperwidth}{\centering\footnotesize
        This work has been submitted to the IEEE for possible publication. Copyright may be transferred without notice, after which this version may no longer be accessible.}}%
      }%
    }%
  }%
}

\begin{document}

\markboth{Submitted to IEEE Transactions on Pattern Analysis and Machine Intelligence}%
{Wang \MakeLowercase{\textit{et al.}}: IndustrialKnowEval: A Benchmark for Evaluating Multi-Level Industrial Knowledge in LLMs}

\title{IndustrialKnowEval: A Benchmark for Evaluating Multi-Level Industrial Knowledge in LLMs}

\author{
{
Haiteng Wang~\orcidlink{0000-0002-7316-3607},~\IEEEmembership{Member,~IEEE,}
Haoze Li~\orcidlink{0009-0006-1834-0048},
Jingwei Guo~\orcidlink{0009-0009-0795-541X},~\IEEEmembership{Student Member,~IEEE,}
Jingbo Zhou~\orcidlink{0000-0003-2677-7021},~\IEEEmembership{Member,~IEEE,}
Si Liu~\orcidlink{0000-0002-9180-2935},~\IEEEmembership{Member,~IEEE,}
and Lei Ren~\orcidlink{0000-0001-6346-6930},~\IEEEmembership{Senior Member,~IEEE}
}

\thanks{
    {	
Haiteng Wang and Si Liu are with the School of Artificial Intelligence, Beihang University, Beijing 100191, China (email: wanghaiteng@buaa.edu.cn, liusi@buaa.edu.cn).

Haoze Li is with the National College for Excellent Engineers, Beihang University, Beijing 100191, China (email: lihaozebuaa@buaa.edu.cn).

Jingwei Guo is with the School of Automation Science and Electrical Engineering, Beihang University, Beijing 100191, China (email: BeihangGjw2403@buaa.edu.cn).

Jingbo Zhou is with the Business Intelligence Lab, Baidu Research, Beijing 100085, China (email: zhoujingbo@baidu.com).

Lei Ren is with the School of Automation Science and Electrical Engineering, Beihang University, Beijing 100191, China, also with the Hangzhou International Innovation Institute, Beihang University, Hangzhou 311115, China, and also with the State Key Laboratory of Intelligent Manufacturing System Technology, Beijing 100854, China (e-mail: renlei@buaa.edu.cn).
    }
}
}

\arxivnotice
\maketitle
\thispagestyle{empty}

\begin{abstract}
\rev{
Large language models (LLMs) are increasingly explored for industrial tasks, yet few benchmarks evaluate their mastery of industrial knowledge across different types of cognitive demand. We introduce IndustrialKnowEval (IndKnow), a benchmark for industrial knowledge and engineering question answering. It contains 43,282 samples covering 21 task types across seven industrial sectors, including aerospace, electric power, and metallurgy, organized into four levels of cognitive demand: memory, comprehension, reasoning, and engineering application. Enterprise-based domain specialists and experienced researchers participate in all stages of dataset construction and quality control. We conduct a series of experiments on 18 representative LLMs, with domain experts deeply involved in the evaluation. Building upon IndKnow, we propose the Industrial Cognitive Capability Gap (ICCG) metric to characterize the performance differential between foundational industrial knowledge and higher-order reasoning and application, and we analyze it alongside absolute performance to provide a more complete view of model capabilities. We anticipate that IndustrialKnowEval will support the benchmarking of LLMs in industrial scenarios and promote the development of robust industrial intelligence.  Code is available at:
\url{https://github.com/J4cksonLee/IndKnow}.
}
\end{abstract}

\rev{
}

\section{Introduction}
The rapid advancement of large language models (LLMs) has significantly enhanced core capabilities in natural language processing ~\cite{brown2020language, cheng2026prompt,yang2025qwen3}, knowledge reasoning ~\cite{yao2023tree, 11123142}, and cross-modal understanding ~\cite{10445007,10876763, wang2024uni}. These technological breakthroughs have not only accelerated the progress of artificial general intelligence but have also demonstrated strong transferability across a wide range of vertical domains. In the industrial sector in particular, LLMs have been increasingly explored for various applications, including equipment condition monitoring, intelligent fault diagnosis, predictive maintenance, production scheduling and process optimization, as well as industrial document analysis. These applications highlight the considerable potential of LLMs to improve production efficiency, reduce operational and maintenance costs, and enhance complex decision-making capabilities.

\rev{We need a benchmark that systematically tests the industrial knowledge of LLMs, so that we can use these models reliably in industrial environments. Existing evaluation frameworks do not yet meet this need. MMLU~\cite{hendrycks2020measuring} and C-Eval~\cite{huang2023c} are general-purpose benchmarks. They include engineering questions, but these questions make up less than 5\% of the total, and most of them test factual recall (Bloom's Level 1--2). Researchers have also built specialized industrial benchmarks such as MMAD~\cite{jiang2024mmad}, MME-Industry~\cite{yi2025mme}, and FDM-Bench~\cite{eslaminia2024fdm}, but these benchmarks cover only a narrow scope. We see two main limitations in existing work:}

\textbf{Limited Evaluation of Higher-Order Engineering Reasoning and Application.} \rev{Industrial knowledge includes more than factual definitions (Know-what). It also includes procedural workflows and physical mechanisms (Know-how). General-purpose benchmarks such as MMLU and C-Eval include only a few industry-related tasks, and these tasks mainly test factual recall or conceptual understanding. These benchmarks therefore tell us little about how well a model can reason with industrial knowledge.}

\textbf{Insufficient Cross-Industry and Cross-Task Coverage.} \rev{Specialized industrial benchmarks focus on a single task or sector. They do not offer a unified design that covers multiple cognitive levels and industries. Aerospace and metallurgy share basic engineering principles, but they differ widely in technical specifications and safety-critical requirements. A benchmark with a narrow scope gives an incomplete evaluation, so it cannot show how well a model understands industrial knowledge as a whole.}

\begin{figure*}[t]
    \centering
    \includegraphics[width=0.7\linewidth]{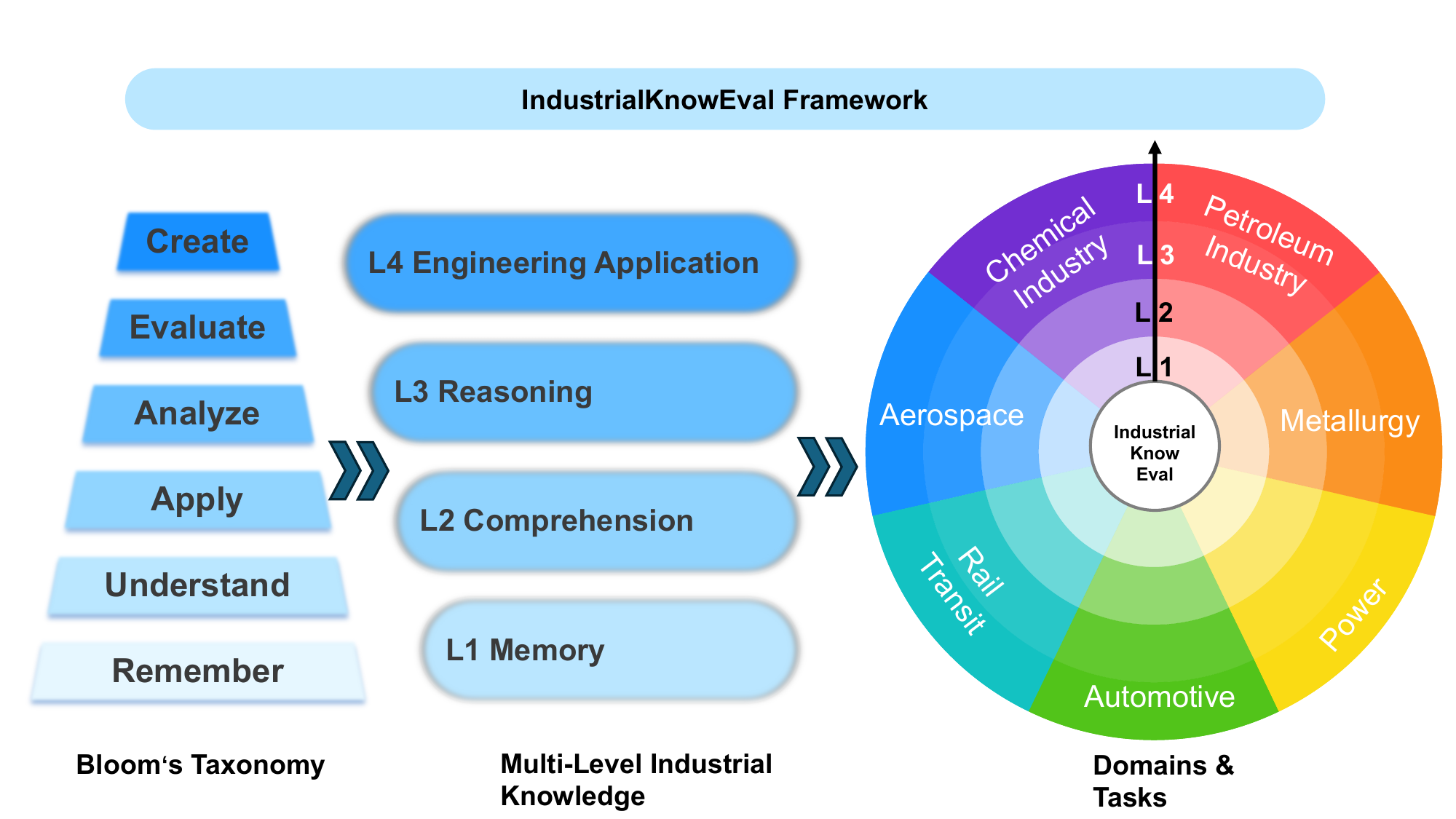}
    \caption{
    \rev{Overview of the IndustrialKnowEval Framework. The framework constructs a four-level cognitive hierarchy (L1–L4) grounded in Bloom’s Taxonomy, mapping cognitive processes to types of industrial questions, from L1 Memory to L4 Engineering Application. The circular schema on the right illustrates the multi-dimensional evaluation matrix, applying this stratified standard across seven core industrial domains to enable a comprehensive assessment of model knowledge.}
    }
    \label{fig:industrialknoweval}
\end{figure*}
\rev{To address these shortcomings, we propose IndustrialKnowEval (IndKnow), a hierarchical evaluation framework for industrial knowledge and engineering question answering. Inspired by the human learning process, IndustrialKnowEval establishes a four-level evaluation framework including memory, comprehension, reasoning, and engineering application. This structure is designed to characterize a model's industrial knowledge across different types of cognitive demand. IndKnow consists of 43,282 question--answer pairs, systematically organized into 21 task types across 7 major industrial sectors---including aerospace, electric power, metallurgy, rail transit, automotive, petroleum and chemical.}
A key design principle of IndustrialKnowEval is to use human expert judgment at the stages where automated methods fall short. Fifteen researchers with over five years of industrial research experience and seven domain specialists from enterprises took part in the construction pipeline. They worked on template design, quality validation, and error analysis. Each specialist was responsible for one industrial sector.
\rev{Furthermore, we develop an evaluation system that integrates accuracy, Pass@k, and the proposed Industrial Cognitive Capability Gap(ICCG) metric. Using our dataset, we conduct zero-shot evaluations on 18 widely used models, together with chain-of-thought-enhanced evaluations, and compare the strengths and weaknesses of different models on the benchmark. Our contributions are summarized as follows:}

\begin{itemize}

    \item \rev{We propose a multi-level evaluation framework for the industrial knowledge of LLMs. The framework consists of four levels and 21 task types, covering memory, comprehension, reasoning, and engineering application, enabling a systematic characterization of models' industrial knowledge across different types of cognitive demand.}
    
    \item \rev{We introduce IndustrialKnowEval, a large-scale, multi-industry benchmark comprising 43,282 test samples across 7 representative industrial domains and 21 task types, generated from industrial documents with systematic quality control and source traceability.}
    
    \item \rev{We propose the Industrial Cognitive Capability Gap (ICCG) metric and conduct comprehensive large-scale evaluations of 18 representative LLMs on IndustrialKnowEval, revealing their performance differences across tasks and offering guidance for future model improvement.}
\end{itemize}

\section{Related Work}

Evaluating the capabilities of large language models in industrial and engineering domains is essential for advancing intelligent manufacturing and autonomous production systems. Existing evaluation efforts can be grouped into three lines: general-purpose benchmarks, domain-specific industrial benchmarks, and Bloom’s Taxonomy in LLM Benchmarks. Each line of work has made important contributions, but none of them fully meets the need for systematic, multi-level industrial evaluation. Our work aims to fill this gap.

\subsection{General-Purpose LLM Benchmarks}

\rev{Early work on LLM evaluation tested broad knowledge across many disciplines. MMLU~\cite{hendrycks2020measuring} set up the paradigm for multitask evaluation. It contains 57 subjects and over 15,000 questions in STEM, the humanities, and the social sciences. Researchers now widely use it to measure zero-shot and few-shot performance. C-Eval~\cite{huang2023c} and CMMLU~\cite{li2024cmmlu} brought this paradigm to Chinese contexts, and they include questions on engineering, physics, and professional certification. BIG-bench~\cite{srivastava2023beyond} and HELM~\cite{liang2022holistic} widened the scope and metrics of evaluation. MMLU-Pro~\cite{wang2024mmlupro} added more answer choices and raised the share of questions that require reasoning. GPQA~\cite{rein2023gpqa} provides graduate-level questions written by experts, and its authors designed these questions to resist web lookup.}

\begin{table*}[t]
\centering
\caption{Comparison between IndustrialKnowEval and representative evaluation benchmarks.}
\label{tab:benchmark-comparison}
\resizebox{\textwidth}{!}{%
\begin{tabular}{llccccccccl}
\toprule
\multirow{2}{*}{Benchmark} & \multirow{2}{*}{Size} & \multirow{2}{*}{Knowledge Source} & \multicolumn{2}{c}{Task Type} & \multicolumn{4}{c}{Ability} & \multirow{2}{*}{Ind. Domain} \\
\cmidrule(lr){4-5} \cmidrule(lr){6-9}
 &  &  & MCQ & OEQ & L1 & L2 & L3 & L4 &  \\
\midrule

\multicolumn{10}{l}{\textit{General-Purpose Benchmarks}} \\
MMLU~\cite{hendrycks2020measuring}       & 15.9K & Web corpus                       & $\checkmark$ &              & $\checkmark$ &              &              &              & 1 (Elec. Eng.) \\
MMLU-Pro~\cite{wang2024mmlupro}          & 12K   & Web corpus                       & $\checkmark$ &              &              &              & $\checkmark$ &              & 1 (Engineering) \\
C-Eval~\cite{huang2023c}                 & 13.9K & Exam + Textbook                  & $\checkmark$ &              & $\checkmark$ & $\checkmark$ &              &              & 3 (Engineering) \\
GPQA~\cite{rein2023gpqa}                 & 0.4K  & Expert-curated                   & $\checkmark$ &              &              &              & $\checkmark$ &              & 0 \\

\midrule
\multicolumn{10}{l}{\textit{Cognitive-Stratified Benchmarks}} \\
SciEval~\cite{sun2024scieval}            & 18K   & Academic paper + Textbook        & $\checkmark$ & $\checkmark$ & $\checkmark$ &              & $\checkmark$ & $\checkmark$ & 0 \\
SciKnowEval~\cite{feng2025sciknoweval}   & 28K   & Academic paper + Textbook        & $\checkmark$ & $\checkmark$ & $\checkmark$ & $\checkmark$ & $\checkmark$ & $\checkmark$ & 0 \\
MedBloomEval~\cite{zhou2025medbloomeval} & --    & Existing dataset                 & $\checkmark$ &              & $\checkmark$ &              & $\checkmark$ & $\checkmark$ & 0 \\
BloomAPR~\cite{ma2025bloomapr}           & --    & Code corpus                      &              & $\checkmark$ & $\checkmark$ & $\checkmark$ & $\checkmark$ &              & 0 \\

\midrule
\multicolumn{10}{l}{\textit{Domain-Specific Industrial Benchmarks}} \\
MMAD~\cite{jiang2024mmad}                & 39.7K & Existing dataset                 & $\checkmark$ &              & $\checkmark$ &              &              & $\checkmark$ & 1 (Anomaly Det.) \\
FDM-Bench~\cite{eslaminia2024fdm}        & --    & Domain doc.                      & $\checkmark$ & $\checkmark$ & $\checkmark$ &              &              & $\checkmark$ & 1 (Additive Mfg.) \\
ChemBench~\cite{mirza2024chembench}      & 2.7K  & Academic paper + Textbook        & $\checkmark$ & $\checkmark$ & $\checkmark$ &              & $\checkmark$ &              & 1 (Chem.) \\
ElecBench~\cite{zhou2024elecbench}       & --    & Domain doc.                      & $\checkmark$ & $\checkmark$ & $\checkmark$ &              & $\checkmark$ & $\checkmark$ & 1 (Power) \\

\midrule
\textbf{Ours (IndustrialKnowEval)} & \textbf{43K} & \textbf{Industrial standards + Academic paper + Textbook } & $\checkmark$ & $\checkmark$ & $\checkmark$ & $\checkmark$ & $\checkmark$ & $\checkmark$ & \makecell[l]{\textbf{7 (Aerospace, Power, Chemical,}\\ \textbf{Metallurgy, Petroleum, Rail,}\\ \textbf{Automotive)}} \\
\bottomrule
\end{tabular}%
}
\end{table*}

\subsection{Domain-Specific Industrial Benchmarks}

\rev{General benchmarks have clear limitations, so researchers have built evaluation frameworks for industrial contexts. MME-Industry~\cite{yi2025mme} is the first cross-industry multimodal benchmark, and it uses image-text questions to cover 21 sectors. MMAD~\cite{jiang2024mmad} focuses on industrial anomaly detection and provides a multimodal dataset of 39K questions. FDM-Bench~\cite{eslaminia2024fdm} targets additive manufacturing and tests user Q\&A and G-code anomaly detection. ElecBench~\cite{zhou2024elecbench} focuses on power dispatch. ChemBench~\cite{mirza2024chembench} collects over 2,700 question-answer pairs in chemistry and materials, and experts designed these pairs. Other benchmarks cover broader engineering disciplines. EngiBench~\cite{felten2025engibenchframeworkdatadrivenengineering} uses a hierarchical structure with three levels, and ENGDESIGN~\cite{guo2025engdesign} combines generative outputs with simulation-based validation.}

\subsection{Bloom’s Taxonomy in LLM Benchmarks}
\rev{Several researchers apply Bloom's Taxonomy~\cite{krathwohl2002revision} to benchmark design. They aim to test higher-order capabilities (analyzing, evaluating, creating) and not only lower-order skills (remembering, understanding). SciEval~\cite{sun2024scieval} turned this framework into four evaluation dimensions for the scientific domain. SciKnowEval~\cite{feng2025sciknoweval} extended this work with a five-level hierarchy of knowledge memory, comprehension, reasoning, discernment, and application. It also treats safety as a separate evaluation dimension. MedBloomEval~\cite{zhou2025medbloomeval} measures LLM reasoning across Bloom's cognitive levels in clinical medicine. BloomAPR~\cite{ma2025bloomapr} applies Bloom's Taxonomy to evaluate LLM-powered automated program repair. Engineering expertise has a different cognitive structure, which includes foundational technical knowledge, procedural skills, and cross-system process reasoning. To our knowledge, few benchmarks have systematically adapted cognitive levels to this structure.}

\rev{As shown in Table~\ref{tab:benchmark-comparison}, existing benchmarks are limited in ability coverage, data scale, or task diversity. While cognitive stratification has been applied in scientific and professional domains, to our knowledge no existing work combines broad industrial domain coverage with systematic cognitive-level differentiation. This work develops a large-scale, multi-level industrial knowledge evaluation framework that spans diverse industrial domains and organizes tasks according to an industrially adapted cognitive hierarchy.}

\section{Methodology}\label{sec:cognitive}

\rev{Based on Bloom's Taxonomy, we organize the capabilities of LLMs in processing industrial knowledge into four levels of cognitive demand, from lower-order to higher-order cognition: Memory, Comprehension, Reasoning, and Engineering Application. These levels range from basic knowledge recall to the application of industrial knowledge to engineering problems described in text.}

\textbf{(1) Memory.} The Memory level tests whether a model can identify and recall key concepts, definitions, standard parameters, technical terms, and basic facts in industrial domains. It measures the breadth and consistency of the model's knowledge across industrial sub-domains.

\textbf{(2) Comprehension.} The Comprehension level tests whether a model understands industrial concepts. Beyond recognizing facts, the model must explain, compare, and synthesize concepts, which requires it to relate pieces of knowledge to one another and describe them coherently and accurately.

\textbf{(3) Reasoning.} \rev{The Reasoning level tests logical inference and analytical problem-solving based on industrial knowledge. Tasks typically involve multi-step reasoning, conditional judgment, or the integration of several pieces of knowledge, and the model must reach a conclusion through analysis.}

\textbf{(4) Engineering Application.} \rev{The Engineering Application level tests whether a model can apply industrial knowledge to engineering problems, including root-cause analysis, fault diagnosis, condition-impact reasoning, consequence prediction, and system optimization. Tasks are based on engineering contexts derived from industrial documents, and the model must give technical solutions, optimization strategies, or engineering recommendations in free text.}

\rev{The four levels represent different types of cognitive demand. They test a model's ability to recall, understand, integrate, and apply industrial knowledge. We do not order these levels by difficulty, because the difficulty of an item for a model depends on the specific knowledge it tests. Level-wise scores therefore describe a model's profile across these types of demand.}

\rev{IndustrialKnowEval uses these four levels as a common scheme, so we can describe and compare different LLMs in the same way.}

\section{Dataset Construction}
This section details the data construction pipeline employed in this study, which comprises three main stages: raw data collecting and preprocessing, evaluation task construction, and quality control.

\begin{figure*}[t]
    \centering
    \includegraphics[width=\linewidth]{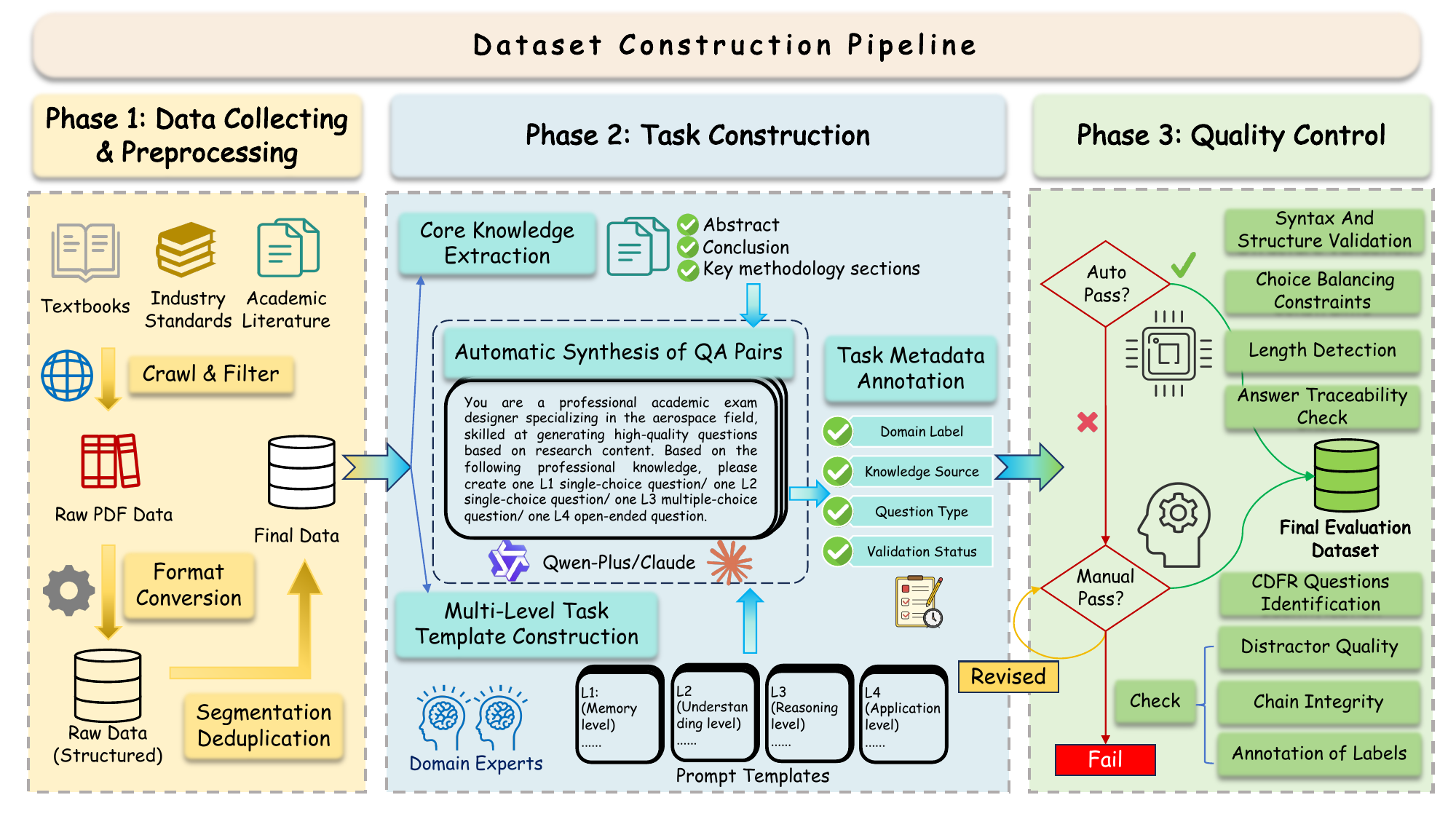}
    \caption{
    An illustration of dataset construction pipeline in IndustrialKnowEval, including I) Raw Data Collecting and Preprocessing, II) Task Construction and III) Quality Control.
    }
    \label{fig:dataset construction}
\end{figure*}

\subsection{Raw Data Collecting and Preprocessing}
\rev{This study focuses on seven representative industrial sectors: aerospace, petroleum, chemical, automotive, metallurgy, rail transit, and electric power. We collected academic literature, industry standards, technical specifications, and textbooks from many sources, so that the benchmark covers a wide range of real-world industrial scenarios. Academic papers came mainly from open-access preprint platforms such as arXiv and from other academic databases. Industry standards and specifications came from documents published by government agencies and industry associations, and textbook material came from open-access higher education engineering courses.}

\rev{We first screened the collected documents by industry theme and removed duplicates at a coarse level. The open-source tool MinerU then converted the remaining PDF files to Markdown in batches. The converted text still contained noise, including references, advertisements, garbled characters, images, and tables. We removed this noise with regular expressions and custom scripts, and the same scripts corrected formatting errors from OCR. Finally, the text was split according to its heading structure. We then removed duplicates again with a keyword-based process and converted all the text to a uniform format.}

\subsection{Task Construction}
\paragraph{Knowledge Point Extraction}
\rev{A rule-based script first split each document into sections according to its Markdown heading hierarchy. An LLM then read the document and extracted its key knowledge fragments. As a result, each document was turned into a set of knowledge fragments with clear meaning and traceable sources. These fragments contain the most informative and knowledge-dense content of the document, including principal conclusions, key process steps, and important parameters.}

\paragraph{Multi-Level Task Template Construction}
We built the multi-level prompt templates in close collaboration with domain experts, so the design rests on authentic industrial knowledge and not only on automated procedures. Seven technical specialists from enterprises took part in this work, and each of them represented a different industrial sector. Fifteen researchers with over five years of industrial research experience also took part. Together, they defined the \rev{cognitive expectations and boundaries} for each evaluation level.
Based on the four-level capability framework of IndustrialKnowEval, we established a mapping between task types and capability levels. We also developed fine-grained prompt templates for each level. These templates specify the authoring role, question formats, general rules, and additional constraints for both multiple-choice and open-ended questions. All templates strictly require that answers rely only on the provided professional knowledge. The templates also prohibit the introduction of external information.
This expert-driven calibration is essential and non-substitutable for ensuring the validity and rigor of the evaluation. The complete templates are provided in the Appendix~\ref{prompt}.

\paragraph{Automatic Synthesis of QA Pairs}
\rev{Based on the extracted knowledge fragments and the constructed template constraints, we used a template-based method to synthesize question-answer (QA) pairs automatically. We embedded each knowledge fragment into the template input as context and synthesized one question per cognitive level (L1--L4) from it. All questions were synthesized by Qwen-Plus. The model followed the template instructions and constraints, and it produced each question together with its standard answer and a detailed explanation.}

\paragraph{Task Metadata Annotation}
\rev{Each question was annotated with the following metadata to support evaluation and later analysis.}

Domain Label: Each question is labeled with one of seven industrial sectors according to its source and content: aerospace, petroleum, chemical, automotive, metallurgy, rail transit, or electric power.

Knowledge Source: Each question is annotated with its source document and location, including the document ID and paragraph index, so that it can be traced to the original text.

Question Type: This label indicates the cognitive objective of the question and how it requires knowledge to be used. Domain experts first analyzed the tasks at each level and defined a set of question types with clear boundaries. The research team then designed prompt templates based on these types and used a state-of-the-art model to assign the labels automatically. The definitions are given in Appendix~\ref{prompt}.

\rev{Validation Status: This field records whether a question passed the automated and manual verification. It takes one of five values, which are defined in Appendix~\ref{qualitycontrol}.}

\subsection{Quality Control}
\rev{Each generated question passed through automated verification and then manual review.}

\paragraph{Automated Verification}
\rev{A script first checked the format of every generated question, including the question stem, the number of options, and the answer key. Choice balancing constraints were then applied to the options of the choice questions, and open-ended questions with very short answers were removed. Finally, we checked traceability: using the source identifier of each question, we combined semantic similarity matching with reverse keyword retrieval to confirm that the reference answer could be located in, or inferred from, the source fragment.}

\paragraph{Manual Review}
\rev{Questions that passed automated verification were reviewed manually for four aspects. Reviewers first checked context dependence. Some questions can only be answered with access to the source document, so they test retrieval rather than domain knowledge. We call these Context-Dependent Factual Reasoning (CDFR) questions. Two domain experts with industrial research experience checked each question independently against three criteria (Appendix~\ref{qualitycontrol}). We excluded a question if either annotator found that it met any of the criteria. This step excluded 745 questions. For choice questions, reviewers checked that the distractors were relevant to the question and similar in structure to the correct answer. For open-ended questions, they checked that the question, the reasoning, and the answer were logically consistent. Questions with logical gaps, ambiguity, or missing reasoning steps were marked \texttt{manual\_fail}. Reviewers then checked the automatically assigned task types and corrected those that were ambiguous or inconsistent. Questions that failed the review were revised or discarded, and a \texttt{validation\_status} field recorded the result of the review.}

\subsection{\rev{Data Validity Audits}}\label{sec:audits}
\rev{\textbf{Filtering statistics.} We synthesized one question per level from each of the 12,151 source texts. This step gave an initial pool of 48,604 questions. Automated verification removed 3,100 questions. Manual review removed 2,222 questions, including 745 CDFR exclusions and 1,477 questions that failed the other review criteria. The final benchmark contains the remaining 43,282 questions (Appendix~\ref{qualitycontrol}).}

\rev{\textbf{Contamination.} The textbooks and standards among the source documents are older, and most research papers were published roughly in 2023--2025. All source documents are public. We measured n-gram overlap between the 188,719 non-empty L4 responses of the 18 models and the source passage of each item, using character 8-grams for Chinese and word 5-grams for English and excluding n-grams that also occur in the question. On average, 0.09\% of the n-grams of a response occur in its source passage, compared with less than 0.001\% for the passage of another item and 9.6\% for the reference answers, which were derived from these passages. Only 100 responses (0.05\%) reach 10\% overlap, three reach 25\%, and none reach 50\%. With character 4-grams and word 3-grams, the means are 0.72\% for the source passage and 0.05\% for another passage. }

\rev{\textbf{Generator preference.} We generated an additional 300 L1--L3 items with Claude Opus 5.5 from 100 source texts of the seven domains, using the same input and template. We evaluated 12 models on both versions (Appendix~\ref{app:generator}). The model ranking is consistent across generators (Spearman $\rho=0.79$, 95\% CI [0.37, 0.91]), with GPT-6-sol highest and Llama-3.1-8B lowest under both. Models of the Qwen family score 0.022 lower on Qwen-generated items than on Claude-generated items, compared with 0.005 higher for the other models (difference $-0.026$, 95\% CI [$-0.064$, 0.011]), so we detect no generator-family preference.}

\subsection{Final Dataset}
The final IndustrialKnowEval dataset contains \rev{43,282 questions across four cognitive levels} and seven industrial sectors: aerospace, petroleum, chemical, automotive, metallurgy, rail transit, and electric power. It is designed to cover both the breadth and the complexity of industrial knowledge. Tables~\ref{tab:dataset overview} and~\ref{tab:task types} present the overall statistics.
\\
\\
\begin{table}[h!]
  \centering
  \footnotesize
  \caption{Overview of the dataset.}
  \label{tab:dataset overview}
  \resizebox{\linewidth}{!}{
  \begin{tabular}{c|ccccccc|c}
  \toprule
  Level & Power& Metallurgy & Chemical & Rail Transit & Automotive & Petroleum & Aerospace& Total\\
  \midrule
  L1 & 1754 & 1731 & 1771 & 700 & 1203 & 1846 & 1940 & 10945 \\
  L2 & 1732 & 1740 & 1765 & 708 & 1204 & 1835 & 1935 & 10919 \\
  L3 & 1754 & 1720 & 1779 & 708 & 1205 & 1828 & 1936 & 10930 \\
  L4 & 1632 & 1679 & 1686 & 691 & 1143 & 1738 & 1919 & 10488 \\
  \midrule
  total & 6872 & 6870 & 7001 & 2807 & 4755 & 7247 & 7730 & 43282 \\
  \bottomrule
  \end{tabular}
  }
\end{table}

\begin{table*}[h!]
  \centering
  \caption{Detailed Questions \& Task Types}
  \label{tab:task types}
  \resizebox{\linewidth}{!}{
  \begin{tabular}{c|cccc}
  \toprule
  \textbf{Level} &\textbf{L1} (Single-Choice Questions) & \textbf{L2} (Single-Choice Questions) &\textbf{L3} (Multiple-Answer Questions) &\textbf{L4} (Open-Ended Questions) \\
  \midrule
  \multirow{6}{*}{\makecell{Task Types}}
  &Method Recognition & Principle Explanation & Influence Factor Analysis & Condition-Impact Reasoning\\
  &Data Extraction & Function Description & Method Analysis & System Optimization \\
  &Concept Recall & Structural Feature Description & Performance Comparison & Fault Diagnosis \\
  &Entity Recognition & Process Description & Process Flow Analysis & Root Cause Reasoning \\
  &Category Identification & Formula Comprehension & Phenomenon Interpretation & Consequence Prediction\\
  &- & - & Structure Function Analysis &  - \\ 
  \bottomrule
  \end{tabular}
  }
\end{table*}
\section{Experiment}
\subsection{Experimental Setup}\label{sec:setup}
\subsubsection{Baselines}
This study selected \rev{18} high-performance LLMs. They were categorized into two main groups based on their technical architecture and open-source strategy:

\textbf{Proprietary API-based Models}: These models typically represent the state-of-the-art in language model development and exhibit powerful, all-around performance. For this category, the study selected \rev{GPT-6-sol, GPT-5, GPT-4.1~\cite{openai2025gpt41}, Gemini 2.5 Pro~\cite{comanici2025gemini}, Qwen-Plus(Qwen-Plus-2025-09-11)~\cite{yang2025qwen3}, GLM-4.5, GLM-4.5-Air ~\cite{zeng2025glm}, GLM-4~\cite{glm2024chatglm}, DeepSeek-V4-Pro-0813~\cite{deepseek2026v4} and DeepSeek-V3.1~\cite{liu2024deepseek}}.

\textbf{Open-weight Foundational Models}: Valued for their open research nature and reproducibility, these models are central to academic research and domain-specific fine-tuning. We selected models from three widely used open-weight model families. They were the LLaMA series (Llama-3.1-8B, Llama-3.1-70B)~\cite{grattafiori2024llama}, the open-source GLM series (GLM-4-9B)~\cite{glm2024chatglm}, and the open-source Qwen series (Qwen3-0.6B, Qwen3-1.7B, Qwen3-4B, Qwen3-8B, Qwen3-14B)~\cite{yang2025qwen3}.

All models were accessed through official APIs or deployed using standardized frameworks. Details of model versions and deployment configurations are provided in Appendix~\ref{modeldetails}.

\subsubsection{Evaluation Methodology}
We evaluated the models under two experimental settings. They are zero-shot and Chain-of-Thought (CoT). \rev{We set the temperature parameter to 0 in the zero-shot and CoT experiments to minimize output randomness, unless otherwise noted in Appendix~\ref{modeldetails}.}

We first evaluated all \rev{18} models with a standard zero-shot prompting template. We then used a CoT prompt to a subset of the models. \rev{This setting used 500 randomly sampled questions at each of L3 and L4 in each domain}. This prompt asks the model to write out its detailed intermediate reasoning steps before it gives a final answer. \rev{Besides the prompt, the CoT setting also changed the reasoning settings of some models. It enabled the thinking mode of the Qwen series, DeepSeek-V3.1, and GLM-4.5, which was disabled in the zero-shot setting, and it raised the reasoning effort of GPT-5 and Gemini 2.5 Pro from the lowest level to the default. }

\subsubsection{Metrics}
This study employs a hierarchical set of quantitative metrics tailored to the different task output formats.

\rev{Accuracy: L1 and L2 single-choice questions and L3 multi-answer questions have definitive answers, so we use accuracy as the primary evaluation metric for them. An L3 response counts as correct only when the predicted options exactly match the reference options. A partially correct response receives no credit. We report micro-averaged scores, in which each item has equal weight, and macro-averaged scores, in which each of the 21 task types has equal weight.}

CoT Gain: This derived metric is defined as the difference between a model's performance on any given metric under the Chain-of-Thought (CoT) setting and its performance on the same metric under the zero-shot baseline setting\rev{, paired on the same model and question}.

\rev{Pass@k: This metric is computed for L3 questions in Chinese and English for four models (GPT-5, Qwen-Plus, Qwen3-8B, and Llama-3.1-8B). Unlike the other experiments, this experiment used sampling: with the zero-shot prompt, $n=5$ responses per item were sampled at temperature 1, with top-$p$ and the random seed left at their API defaults; the thinking mode of the Qwen models was disabled, and the reasoning effort of GPT-5 was set to minimal. A sample is correct only if its set of options exactly matches the reference set. Pass@k is the probability that at least one of $k$ samples is correct, estimated as $1-\binom{n-c}{k}/\binom{n}{k}$ for $k=1,\dots,5$, where $c$ is the number of correct samples among the $n$ samples.}

\rev{Score (by LLM-as-a-Judge): Open-ended L4 questions were scored by an LLM judge. Our primary judge was Qwen-Plus (qwen-plus-2025-09-11), which we accessed through the Alibaba Cloud API around November--December 2025, with thinking disabled and the provider's default temperature. It compared each response with the reference answer and gave an integer score $s$ from 1 to 10 under a rubric of accuracy, completeness, relevance, and conciseness (Appendix~\ref{prompt}), and we normalized the scores as $s/10$. We then had all responses re-scored under the same prompt by GPT-6-sol, from a different model family, and by Qwen3.7-Plus. Section~\ref{sec:validity} reports how well the judges agree, along with expert scoring on a sample of responses.}

\rev{Industrial Cognitive Capability Gap (ICCG): To compare a model's performance on higher-order tasks (reasoning and engineering application) with its performance on foundational tasks (memory and comprehension), this study defines ICCG as follows:}
\begin{equation}
\rev{\mathrm{ICCG} = 1 - r, \qquad r = \frac{\overline{\mathrm{Acc}}_{(L3+L4)}}{\overline{\mathrm{Acc}}_{(L1+L2)}}}
\label{eq:iccg}
\end{equation}
\rev{where $\overline{\mathrm{Acc}}_{(L1+L2)}$ is the mean of the L1 and L2 accuracies and $\overline{\mathrm{Acc}}_{(L3+L4)}$ is the mean of the L3 accuracy and the normalized L4 score. ICCG is computed for each domain and then averaged across domains. This value shows how far higher-order performance falls short of foundational performance, as a proportion of foundational performance. A negative value means that higher-order performance exceeds foundational performance. Because the levels are not ordered by difficulty (Section~\ref{sec:cognitive}), ICCG is used to compare models and is not interpreted as a degradation implied by the hierarchy. It is reported alongside absolute performance, and a variant without L4, variants with L4 scored by the two additional judges, and an absolute-difference metric are given in Section~\ref{sec:iccg} and Appendix~\ref{sec:iccg_variants}.}

\rev{We also conduct statistical significance tests on L1--L3 accuracy. Because items derived from the same source document are not independent, confidence intervals and tests are computed with a paired cluster bootstrap~\cite{efron1994introduction} that resamples source documents (11,322 clusters) with 10,000 replicates and a fixed random seed (42). We compare GPT-5 with each of the other 17 models at $\alpha = 0.05$ with Bonferroni correction ($\alpha = 0.05/17 \approx 0.0029$). Detailed results, including those with fragment-level resampling, are provided in Appendix~\ref{sec:stat_test}.}

\subsection{\rev{Main Results}}\label{sec:main_results}

In this section, we report the performance of the large language models on our IndustrialKnowEval dataset. Table \ref{tab:overall performance} summarizes the zero-shot performance of the LLMs in each domain.
\begin{table*}[h]
  \centering
  \footnotesize
  \renewcommand{\arraystretch}{1.3}
  \caption{Overall zero-shot performance of LLMs across four levels in seven domains. \rev{Aerospace refers to the Chinese questions; the level columns, Mean, and Macro also include the English Aerospace questions. Mean is micro-averaged (each item has equal weight), Macro gives each of the 21 task types equal weight, and rank follows Mean. L4 scores are from the primary judge.} \underline{\textbf{Bold underlined}} indicates the best results among all models, \textbf{bold} indicates the second-best results, and \rev{\underline{underline} indicates the third-best results}. Statistical significance tests are reported in Appendix~\ref{sec:stat_test}.}

  \label{tab:overall performance}
  \resizebox{\linewidth}{!}{
  \begin{tabular}{c|ccccccc|cccc|ccc}
  \toprule
  \textbf{Model} & \textbf{Power} & \textbf{Metallurgy} & \textbf{Chemical} & \textbf{Rail Transit} & \textbf{Automotive} & \textbf{Petroleum} & \textbf{Aerospace} & \textbf{L1} & \textbf{L2} & \textbf{L3} & \textbf{L4} & \textbf{Mean} & \rev{\textbf{Macro}} & \textbf{Rank} \\
  \midrule
  \rev{Deepseek-V3.1} & \rev{0.7977} & \rev{0.6591} & \rev{0.7185} & \rev{0.7127} & \rev{0.7609} & \rev{0.7126} & \rev{0.7165} & \rev{0.7481} & \rev{0.8246} & \rev{0.5701} & \rev{0.7649} & \rev{0.7265} & \rev{0.7034} & \rev{7} \\
  \rev{Deepseek-V4-Pro-0813} & \rev{\underline{0.8201}} & \rev{0.6586} & \rev{0.7183} & \rev{0.7079} & \rev{0.7646} & \rev{0.7231} & \rev{0.7128} & \rev{0.7623} & \rev{\underline{0.8439}} & \rev{0.5301} & \rev{\textbf{0.7797}} & \rev{0.7285} & \rev{0.7014} & \rev{6} \\
  \rev{Gemini-2.5-Pro} & \rev{0.8181} & \rev{\underline{0.6855}} & \rev{\underline{0.7295}} & \rev{\underline{0.7294}} & \rev{\underline{0.7784}} & \rev{\underline{0.7344}} & \rev{\underline{0.7425}} & \rev{\underline{0.7731}} & \rev{0.8361} & \rev{\textbf{0.6194}} & \rev{\underline{0.7671}} & \rev{\underline{0.7487}} & \rev{\underline{0.7250}} & \rev{\underline{3}} \\
  \rev{GLM-4.5} & \rev{0.7968} & \rev{0.6480} & \rev{0.7068} & \rev{0.6898} & \rev{0.7509} & \rev{0.7022} & \rev{0.6961} & \rev{0.7444} & \rev{0.8129} & \rev{0.5602} & \rev{0.7565} & \rev{0.7181} & \rev{0.6953} & \rev{8} \\
  \rev{GLM-4.5-Air} & \rev{0.7968} & \rev{0.6421} & \rev{0.7023} & \rev{0.6897} & \rev{0.7399} & \rev{0.6997} & \rev{0.6954} & \rev{0.7386} & \rev{0.8125} & \rev{0.5463} & \rev{0.7495} & \rev{0.7113} & \rev{0.6840} & \rev{9} \\
  \rev{GLM-4} & \rev{0.7499} & \rev{0.5637} & \rev{0.6244} & \rev{0.6143} & \rev{0.6706} & \rev{0.6351} & \rev{0.6100} & \rev{0.6896} & \rev{0.7755} & \rev{0.4444} & \rev{0.6572} & \rev{0.6415} & \rev{0.6087} & \rev{14} \\
  \rev{GLM-4-9B} & \rev{0.7523} & \rev{0.5626} & \rev{0.6251} & \rev{0.6213} & \rev{0.6869} & \rev{0.6355} & \rev{0.6209} & \rev{0.6799} & \rev{0.7631} & \rev{0.5242} & \rev{0.6088} & \rev{0.6443} & \rev{0.6165} & \rev{13} \\
  \rev{GPT-4.1} & \rev{0.8139} & \rev{0.6634} & \rev{0.7235} & \rev{0.7054} & \rev{0.7623} & \rev{0.7183} & \rev{0.7165} & \rev{0.7503} & \rev{0.8302} & \rev{0.6030} & \rev{0.7580} & \rev{0.7351} & \rev{0.7144} & \rev{5} \\
  \rev{GPT-5} & \rev{\textbf{0.8398}} & \rev{\textbf{0.7020}} & \rev{\textbf{0.7572}} & \rev{\textbf{0.7479}} & \rev{\textbf{0.7850}} & \rev{\textbf{0.7538}} & \rev{\textbf{0.7552}} & \rev{\textbf{0.7808}} & \rev{\textbf{0.8527}} & \rev{0.6038} & \rev{\textbf{\underline{0.8433}}} & \rev{\textbf{0.7694}} & \rev{\textbf{0.7463}} & \rev{\textbf{2}} \\
  \rev{GPT-6-sol} & \rev{\textbf{\underline{0.8757}}} & \rev{\textbf{\underline{0.7834}}} & \rev{\textbf{\underline{0.8112}}} & \rev{\textbf{\underline{0.8246}}} & \rev{\textbf{\underline{0.8418}}} & \rev{\textbf{\underline{0.8127}}} & \rev{\textbf{\underline{0.8306}}} & \rev{\textbf{\underline{0.8974}}} & \rev{\textbf{\underline{0.9292}}} & \rev{\textbf{\underline{0.7261}}} & \rev{0.7619} & \rev{\textbf{\underline{0.8293}}} & \rev{\textbf{\underline{0.8106}}} & \rev{\textbf{\underline{1}}} \\
  \rev{Llama-3.1-8B} & \rev{0.5568} & \rev{0.3686} & \rev{0.4376} & \rev{0.4324} & \rev{0.4830} & \rev{0.4581} & \rev{0.4067} & \rev{0.4872} & \rev{0.5394} & \rev{0.4114} & \rev{0.4133} & \rev{0.4633} & \rev{0.4464} & \rev{17} \\
  \rev{Llama-3.1-70B} & \rev{0.7148} & \rev{0.5292} & \rev{0.6002} & \rev{0.5960} & \rev{0.6573} & \rev{0.6081} & \rev{0.6096} & \rev{0.6939} & \rev{0.7831} & \rev{0.4738} & \rev{0.5384} & \rev{0.6232} & \rev{0.5976} & \rev{15} \\
  \rev{Qwen3-0.6B} & \rev{0.5165} & \rev{0.3215} & \rev{0.3828} & \rev{0.3727} & \rev{0.4217} & \rev{0.3824} & \rev{0.3640} & \rev{0.4206} & \rev{0.5082} & \rev{0.1347} & \rev{0.5220} & \rev{0.3950} & \rev{0.3696} & \rev{18} \\
  \rev{Qwen3-1.7B} & \rev{0.6887} & \rev{0.4704} & \rev{0.5386} & \rev{0.5391} & \rev{0.6018} & \rev{0.5486} & \rev{0.5137} & \rev{0.5814} & \rev{0.6716} & \rev{0.3384} & \rev{0.6435} & \rev{0.5578} & \rev{0.5198} & \rev{16} \\
  \rev{Qwen3-4B} & \rev{0.7660} & \rev{0.5860} & \rev{0.6406} & \rev{0.6358} & \rev{0.6961} & \rev{0.6479} & \rev{0.6305} & \rev{0.6953} & \rev{0.7784} & \rev{0.4806} & \rev{0.6748} & \rev{0.6571} & \rev{0.6232} & \rev{12} \\
  \rev{Qwen3-8B} & \rev{0.7753} & \rev{0.5935} & \rev{0.6582} & \rev{0.6519} & \rev{0.7139} & \rev{0.6731} & \rev{0.6550} & \rev{0.7112} & \rev{0.7940} & \rev{0.5169} & \rev{0.6742} & \rev{0.6741} & \rev{0.6438} & \rev{11} \\
  \rev{Qwen3-14B} & \rev{0.7663} & \rev{0.6089} & \rev{0.6693} & \rev{0.6578} & \rev{0.7066} & \rev{0.6709} & \rev{0.6488} & \rev{0.7414} & \rev{0.8150} & \rev{0.4408} & \rev{0.7144} & \rev{0.6775} & \rev{0.6457} & \rev{10} \\
  \rev{Qwen-Plus} & \rev{0.8132} & \rev{0.6681} & \rev{0.7173} & \rev{0.7218} & \rev{0.7653} & \rev{0.7241} & \rev{0.7326} & \rev{0.7723} & \rev{0.8432} & \rev{\underline{0.6052}} & \rev{0.7281} & \rev{0.7373} & \rev{0.7154} & \rev{4} \\
  \midrule
  Mean & \rev{0.7588} & \rev{0.5953} & \rev{0.6534} & \rev{0.6472} & \rev{0.6993} & \rev{0.6578} & \rev{0.6476} & \rev{0.7038} & \rev{0.7785} & \rev{0.5072} & \rev{0.6864} & \rev{0.6688} & \rev{0.6426} & \\
  \bottomrule
  \end{tabular}
  }
\end{table*}

\subsubsection{Model Performance}
\rev{Table~\ref{tab:overall performance} shows a wide spread in overall scores, from 0.3950 (Qwen3-0.6B) to 0.8293 (GPT-6-sol). GPT-6-sol ranks first overall and in every domain, as well as on L1, L2, and L3. It does not lead on L4, where GPT-5 gains the highest score under the primary judge (0.8433 vs.\ 0.7619); the dependence of L4 scores on the judge is examined in Section~\ref{sec:validity}.}

\rev{Below GPT-6-sol, the leading models are closely spaced. On L1--L3 accuracy, paired cluster bootstrap tests that resample source documents (10,000 replicates, $n = 32{,}794$, Bonferroni-corrected $\alpha = 0.05/17 \approx 0.0029$) show no significant difference between GPT-5 and either Gemini-2.5-Pro ($p = 0.21$) or Qwen-Plus ($p = 0.011$). GPT-5 significantly outperforms the other 14 models below it, and GPT-6-sol significantly outperforms GPT-5 (0.8509 vs.\ 0.7458, $p < 0.001$). Small differences in rank among the leading models should therefore not be overinterpreted. Detailed results are provided in Appendix~\ref{sec:stat_test}.}

\rev{Within the open-source Qwen series, overall scores increase with parameter count (0.3950, 0.5578, 0.6571, 0.6741, and 0.6775 from 0.6B to 14B). The gain is large among the smaller models. The score rises by 0.1628 from 0.6B to 1.7B and by 0.0993 from 1.7B to 4B. Beyond 4B, the increase slows, with a gain of 0.0170 from 4B to 8B and only 0.0034 from 8B to 14B.}

\subsubsection{Performance on Each Industry Domain}
\rev{Scores differ across domains. Averaged over the 18 models, Power (0.7588) and Automotive (0.6993) have the highest scores, and Power leads Automotive by 0.0595. Petroleum (0.6578), Chemical (0.6534), Aerospace (0.6476), and Rail (0.6472) have similar scores, and these four domains differ by only 0.0106. Metallurgy has the lowest score (0.5953), which is 0.0519 below Rail and 0.1635 below Power. All 18 models show the same pattern at both ends. Each model scores highest in Power and lowest in Metallurgy. Even GPT-6-sol falls from 0.8757 in Power to 0.7834 in Metallurgy. Its gap of 0.0923 is smaller than the average gap, but a high overall score does not remove the difference between domains.}

\rev{Part of the difference in Metallurgy reflects task composition rather than domain knowledge: Data Extraction, the most error-prone L1 task type, accounts for 43\% of its L1 items, compared with 8\% in Power. We reweighted each domain to a common task-type distribution within each level. After this reweighting, the L1--L3 accuracy gap between Metallurgy and the mean of the other domains falls from 9.1 to 5.4 percentage points. Task composition therefore explains about 40\% of the gap. The remaining gap appears mainly in conceptual and process-related task types (Appendix~\ref{sec:case_study}).}

\rev{An error analysis of all 18 models is given in Appendix~\ref{sec:case_study}. The L3 error patterns differ by family: over-selection dominates for the GPT, GLM, and Qwen families, but under-selection is the most common for the DeepSeek family. Data Extraction at L1 and Formula Comprehension at L2 are the most difficult task types for all models.}

\subsubsection{Performance on Each Level}
\rev{Averaged over the 18 models, the scores are 0.7038 on L1, 0.7785 on L2, 0.5072 on L3, and 0.6864 on L4. Level-wise scores in the full benchmark are jointly determined by the knowledge demands of each level, response format, and scoring protocol: L1 and L2 are single-answer choice questions, L3 multi-answer questions are credited only when the exact set of correct options is selected, and L4 responses are graded by an LLM judge. The lower L3 score and the higher L4 score therefore cannot be attributed to cognitive demand alone. Section~\ref{sec:validity} separates these factors using two controlled subsets in which all levels share the same format and scoring rule, one in single-choice format and one in open-ended format. In these subsets, format and scoring change both the size and the direction of the level-wise differences, and no ordering of the levels holds across formats and grading rules, consistent with the definition of the levels as types of cognitive demand rather than difficulty tiers (Section~\ref{sec:cognitive}).}

\subsubsection{Performance on Different Languages} 
\rev{Averaged over the 18 models, scores are higher in English than in Chinese, and this holds for 16 of the 18 models. The difference varies by level: English scores are higher on L2 for all 18 models and on L3 for 15, but on L1 for only 5. It also varies by model. The largest differences are observed for Llama-3.1-8B (0.5711 vs.\ 0.4067) and Llama-3.1-70B (0.6857 vs.\ 0.6096), followed by GPT-5 (0.8236 vs.\ 0.7552), but Qwen3-8B and DeepSeek-V4-Pro-0813 score slightly lower in English. Model size does not account for this variation: GPT-5 shows a larger difference than the small Qwen3-0.6B (0.3699 vs.\ 0.3640).}

\subsection{\rev{Measurement Validity}}\label{sec:validity}
\rev{The levels in the full benchmark differ in question format and scoring, and this section uses two controlled subsets to separate these two factors from cognitive demand. It then examines the L4 scores with two additional judges and with experts.}

\rev{\textbf{Controlled subsets.} We built two subsets of 350 items per level, sampled at random from the main benchmark within each domain and level, with no source item shared between them. All items are single-choice questions in one subset and open-ended questions in the other. In the single-choice subset, the L1 and L2 items are used as they are. The L3 items were turned into single-answer questions by a fixed rule, with the exact set of correct statements as the correct option, and each L4 item received a condensed reference answer as the correct option and three plausible distractors. In the open-ended subset, the L4 items are used as they are, and Qwen-Plus rewrote the L1 to L3 items as open-ended questions whose reference answers come from the correct options. Each conversion keeps the knowledge point and the reasoning of the original item. Three experts, joined by experts from each domain, reviewed every converted item, and only approved items entered a subset. Appendix~\ref{sec:controlled_subsets} gives the conversion rules and prompts.}

\rev{We evaluated 12 models on the single-choice subset and 10 on the open-ended subset. Qwen-Plus grades the open-ended answers, with GPT-6-sol as a second judge, under two rules. One is the L4 rubric of the main benchmark, which gives partial credit. The other is an all-or-nothing rule that mirrors exact set match: an answer counts as correct only when it covers every key point of the reference answer and contradicts none of them. Three experts confirmed this rule on a pilot of 160 items before it was used. Table~\ref{tab:controlled} summarizes the results.}

\begin{table}[!t]
\centering
\caption{\rev{Mean scores on the controlled subsets (12 models on the single-choice subset and 10 on the open-ended subset). L4 is shown for GPT-5 and Qwen-Plus in the single-choice row and for Qwen-Plus in the all-or-nothing row.}}
\label{tab:controlled}
\scriptsize\setlength{\tabcolsep}{1pt}
\begin{tabular*}{\linewidth}{@{\extracolsep{\fill}}lcccc@{}}
\toprule
\textbf{Subset and grading} & \textbf{L1} & \textbf{L2} & \textbf{L3} & \textbf{L4} \\
\midrule
Single-choice & 0.740 & 0.805 & 0.695 & 0.989, 0.974 \\
Single-choice, L3 in multi-answer form & -- & -- & 0.524 & -- \\
Open-ended, L4 rubric, primary judge & 0.482 & 0.561 & 0.522 & 0.715 \\
Open-ended, L4 rubric, GPT-6-sol & 0.520 & 0.572 & 0.525 & 0.553 \\
Open-ended, all-or-nothing, primary judge & 0.219 & 0.140 & 0.059 & 0.180 \\
Open-ended, all-or-nothing, GPT-6-sol & 0.356 & 0.287 & 0.141 & -- \\
\bottomrule
\end{tabular*}
\end{table}

\rev{In the single-choice subset, L2 beats L1 for 11 of the 12 models (mean difference 0.065, 95\% CI [0.026, 0.105]), and L3 falls below L2 for all 12. With every level in the same single-answer format, the mean score rises from 0.740 on L1 to 0.805 on L2 and falls to 0.695 on L3. Models answer comprehension items more accurately than memory items, and integration items less accurately than both. The L3 gap appears for every model, from GPT-6-sol (0.897 on L2 and 0.820 on L3) to Llama-3.1-8B (0.526 and 0.374), and it does not depend on model size or family. Asking the same L3 items in their original multi-answer form lowers the mean further to 0.524, a significant change for 11 models (exact McNemar test with Holm correction). Exact set matching thus adds to an L3 gap that is already present when all levels share one format.}

\rev{The open-ended subset reacts more strongly to grading. Under the L4 rubric, L4 scores highest for all 10 models with the primary judge, but most of that lead disappears with GPT-6-sol, where L4 beats L2 for only 3 of 9 models. The two judges agree on L1--L3 (ICC 0.70--0.89) and part mainly on L4 (mean difference 0.161, ICC 0.65). All-or-nothing grading gives yet another order, with L1 above L2 and L2 above L3 for all 10 models under both judges. Because reference answers grow longer with level and this rule needs every key point, the decline follows the number of key points required.}

\rev{With format and scoring held fixed, L3 scores below L2 in 51 of the 52 model--setting comparisons: for all 12 models in the single-choice subset, for all 10 models under the L4 rubric with the primary judge and for 9 of 10 with GPT-6-sol, and for all 10 models under all-or-nothing grading with both judges. The integration demand of L3 therefore lowers performance across formats and grading rules. The order of L1 and L2 responds to the grading rule, which fits the definition of the levels as distinct types of cognitive demand, and the level-wise scores give each model a profile across these demands. Task types add detail to this profile. Data extraction, the L1 task type that asks for specific values reported in the source, reaches 0.538 on the single-choice subset, below every L2 and L3 question type (0.641 or higher), and marks precise numerical knowledge as a weak point shared by the models.}

\rev{\textbf{L3 scoring.} From the same responses we computed three L3 scores. Averaged over the 18 models, L3 reaches 0.507 under exact match, 0.853 under option-level F1, and 0.784 under the Jaccard index. With F1 it exceeds L2 (mean 0.779) for 17 models, and with the Jaccard index for 9. Each knowledge fragment yielded one question per level, which also lets us hold the source passage fixed. On the 9,554 fragments that kept questions at all four levels (84.2\% of all fragments), the scores are 0.706 on L1, 0.780 on L2, 0.508 on L3, and 0.689, 0.545, and 0.551 on L4 under the primary, second, and third judges. These values are close to those of the full benchmark, and the level-wise pattern does not come from differences in source composition.}

\rev{\textbf{Cross-judge evaluation.} GPT-6-sol and Qwen3.7-Plus re-scored all 188,764 L4 responses (Table~\ref{tab:judges}). The primary judge scores 0.140 higher than GPT-6-sol (95\% CI [0.139, 0.141]) and 0.138 higher than Qwen3.7-Plus (95\% CI [0.137, 0.140]), and the two additional judges agree closely with each other ($\rho=0.891$, ICC 0.888, mean difference 0.002). L4 exceeds L3 for all 18 models under the primary judge and for 13 and 12 models under the other two. Under the additional judges, no model scores higher on L4 than on L1 and L2. The size of the L4 advantage thus varies with the judge, and the primary judge is the most lenient of the three. The judges still rank the models in nearly the same order (Kendall $\tau=0.843$ and 0.869 against the primary judge).}

\begin{table}[!t]
\centering
\caption{\rev{Agreement of L4 scores among the LLM judges over all L4 responses. Mean difference is the first judge minus the second.}}
\label{tab:judges}
\scriptsize\setlength{\tabcolsep}{1pt}
\begin{tabular*}{\linewidth}{@{\extracolsep{\fill}}lcccc@{}}
\toprule
\textbf{Comparison} & \textbf{Spearman $\rho$} & \textbf{ICC} & \textbf{Quadratic $\kappa$} & \textbf{Mean diff.} \\
\midrule
Primary vs.\ GPT-6-sol & 0.831 & 0.694 & 0.694 & 0.140 \\
Primary vs.\ Qwen3.7-Plus & 0.893 & 0.734 & 0.734 & 0.138 \\
GPT-6-sol vs.\ Qwen3.7-Plus & 0.891 & 0.888 & 0.888 & $-$0.002 \\
\bottomrule
\end{tabular*}
\end{table}

\rev{Response length and model family have little effect on the scores. With item fixed effects, one standard deviation of log length adds 0.036 to the normalized score under the primary judge and 0.007 under GPT-6-sol. With model fixed effects added, the coefficients fall to 0.004 and $-$0.001, and the association comes from differences between models. Relative to GPT-6-sol, the primary judge rates responses of Qwen models 0.001 higher (95\% CI [$-0.001$, 0.002]), with no sign of a preference for its own family. GPT-6-sol, itself an evaluated model, rates its own responses 0.199 higher than those of GPT-5 and GPT-4.1 relative to the primary judge (95\% CI [0.196, 0.201]), which supports its role as a second judge and not as the primary one.}

\rev{\textbf{Expert scoring.} Three experts scored the same 50 L4 responses independently, without seeing model names or judge scores. The responses were drawn at random to cover all 18 models and all eight domain versions. The experts agree at ICC(A,1) $=0.878$ (95\% CI [0.820, 0.913]), with pairwise Spearman $\rho$ from 0.755 to 0.817. Table~\ref{tab:experts} compares each judge with the mean expert score. The expert mean lies between the judges: the primary judge is 0.088 higher, GPT-6-sol 0.058 lower, and Qwen3.7-Plus 0.020 lower. GPT-6-sol follows the expert ranking most closely ($\rho=0.907$), in line with the cross-judge results. Two checks leave these results unchanged. Without the 12 responses for which an expert noted unfamiliarity with the topic or used machine translation, the mean differences are 0.096, $-0.062$, and $-0.020$. Raising the seven expert scores of 0 to the judges' minimum of 1 moves the statistics by at most 0.005.}

\begin{table}[!t]
\centering
\caption{\rev{L4 judges compared with the mean score of three experts on 50 responses, with 95\% bootstrap confidence intervals. $\rho$ is the Spearman correlation, and mean diff.\ is judge minus experts.}}
\label{tab:experts}
\tiny\setlength{\tabcolsep}{1pt}
\begin{tabular*}{\linewidth}{@{\extracolsep{\fill}}lccc@{}}
\toprule
\textbf{Judge} & \textbf{$\rho$} & \textbf{ICC(A,1)} & \textbf{Mean diff.} \\
\midrule
Primary & 0.756 [0.569, 0.869] & 0.821 [0.734, 0.879] & 0.088 [0.050, 0.127] \\
GPT-6-sol & 0.907 [0.809, 0.956] & 0.938 [0.891, 0.965] & $-$0.058 [$-$0.082, $-$0.035] \\
Qwen3.7-Plus & 0.759 [0.568, 0.882] & 0.899 [0.823, 0.943] & $-$0.020 [$-$0.057, 0.016] \\
\bottomrule
\end{tabular*}
\end{table}

\subsection{CoT Performance on IndustrialKnowEval}\label{sec:cot}
\rev{To evaluate the CoT setting, this study compares it with the zero-shot setting at L3 (Reasoning) and L4 (Engineering Application). Five hundred questions were randomly sampled at each level in each domain, and gains are paired on the same model and question, over the sampled questions retained in the final benchmark. The tested models are DeepSeek-V3.1, Gemini 2.5 Pro, GLM-4.5, GPT-4.1, GPT-5, and the Qwen series (Qwen3-0.6B to Qwen-Plus).} Specific prompts are listed in Appendix~\ref{prompt}.
\begin{figure}[H]
    \centering
    \includegraphics[width=0.6\linewidth]{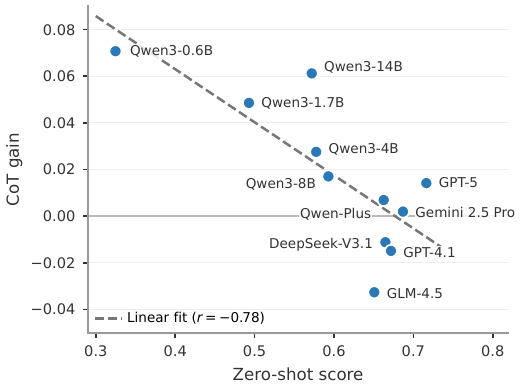}
    \caption{\rev{Zero-shot score and CoT gain for each of the 11 tested models, averaged over L3 and L4 on the Chinese questions of the seven domains. Lower-scoring models tend to obtain larger gains.}}
    \label{fig:inverse-scaling}
\end{figure}

\begin{figure}[H]
    \centering
    \includegraphics[width=0.6\linewidth]{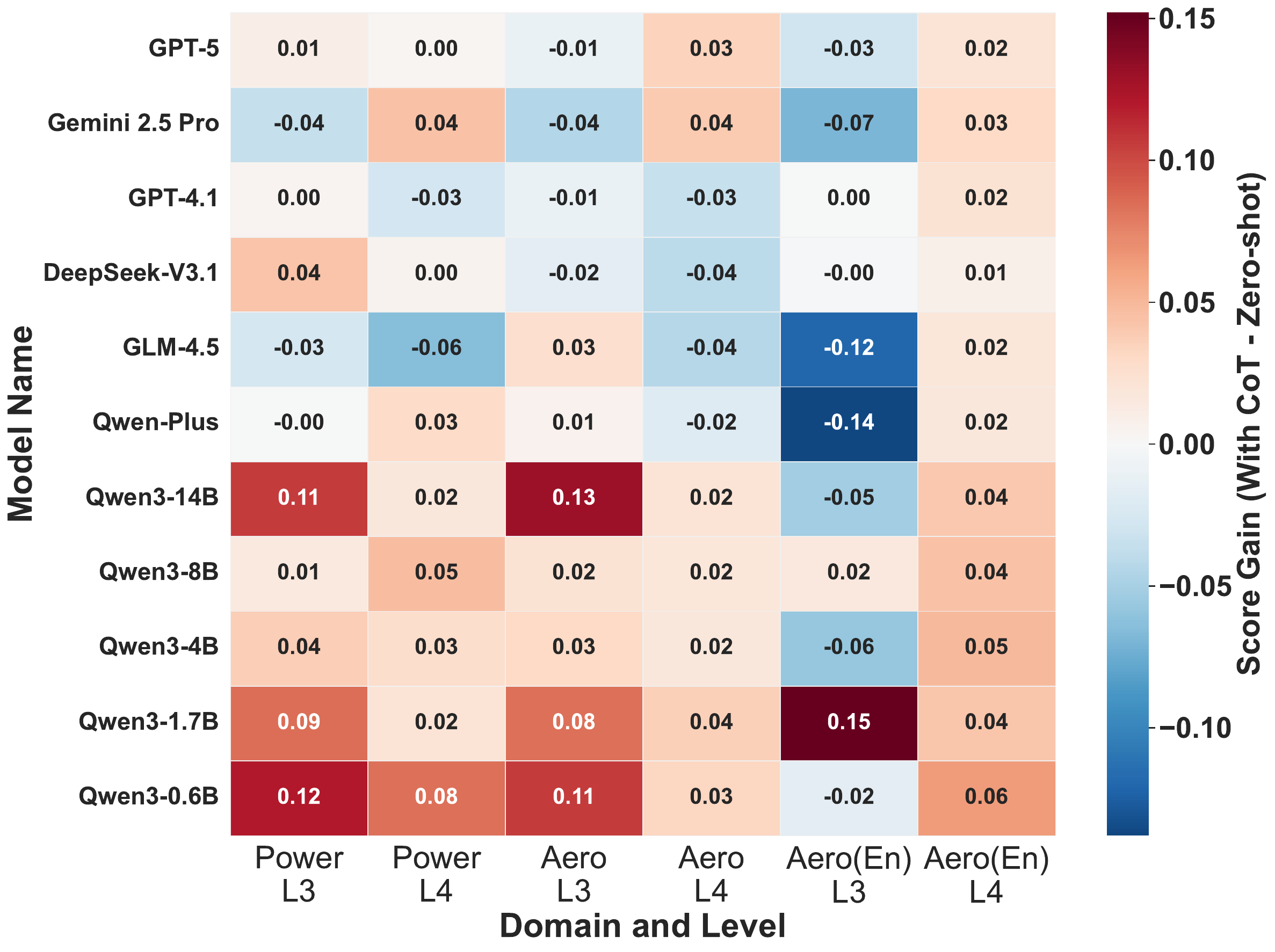}
    \caption{\rev{CoT gains in Power, Aerospace, and Aerospace (English) at L3 and L4.}}
    \label{fig:cot-heatmap}
\end{figure}

\begin{figure}[!t]
    \centering

    \begin{subfigure}{0.45\linewidth}
        \centering
        \includegraphics[width=\linewidth]{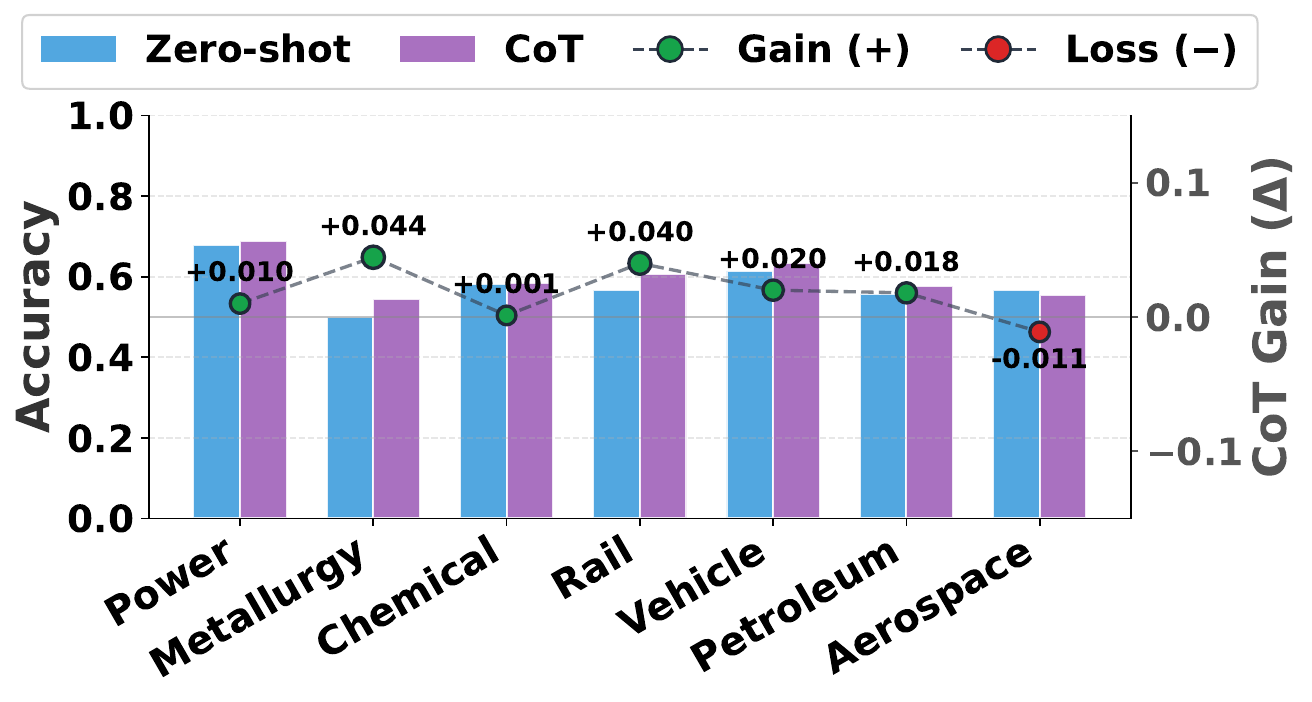}
        \caption{GPT-5 on L3 reasoning tasks.}
        \label{fig:cot_gpt5_l3}
    \end{subfigure}
    \hfill
    \begin{subfigure}{0.45\linewidth}
        \centering
        \includegraphics[width=\linewidth]{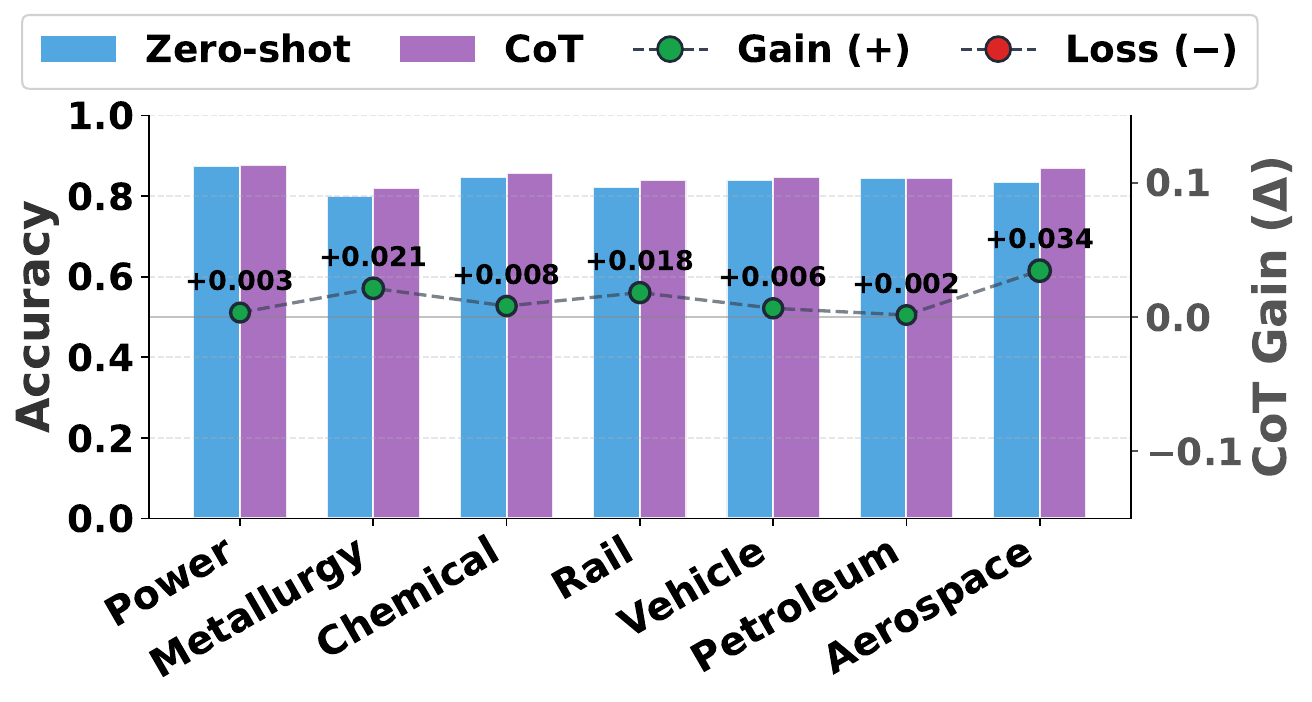}
        \caption{GPT-5 on L4 engineering application tasks.}
        \label{fig:cot_gpt5_l4}
    \end{subfigure}

    \vspace{1em}

    \begin{subfigure}{0.45\linewidth}
        \centering
        \includegraphics[width=\linewidth]{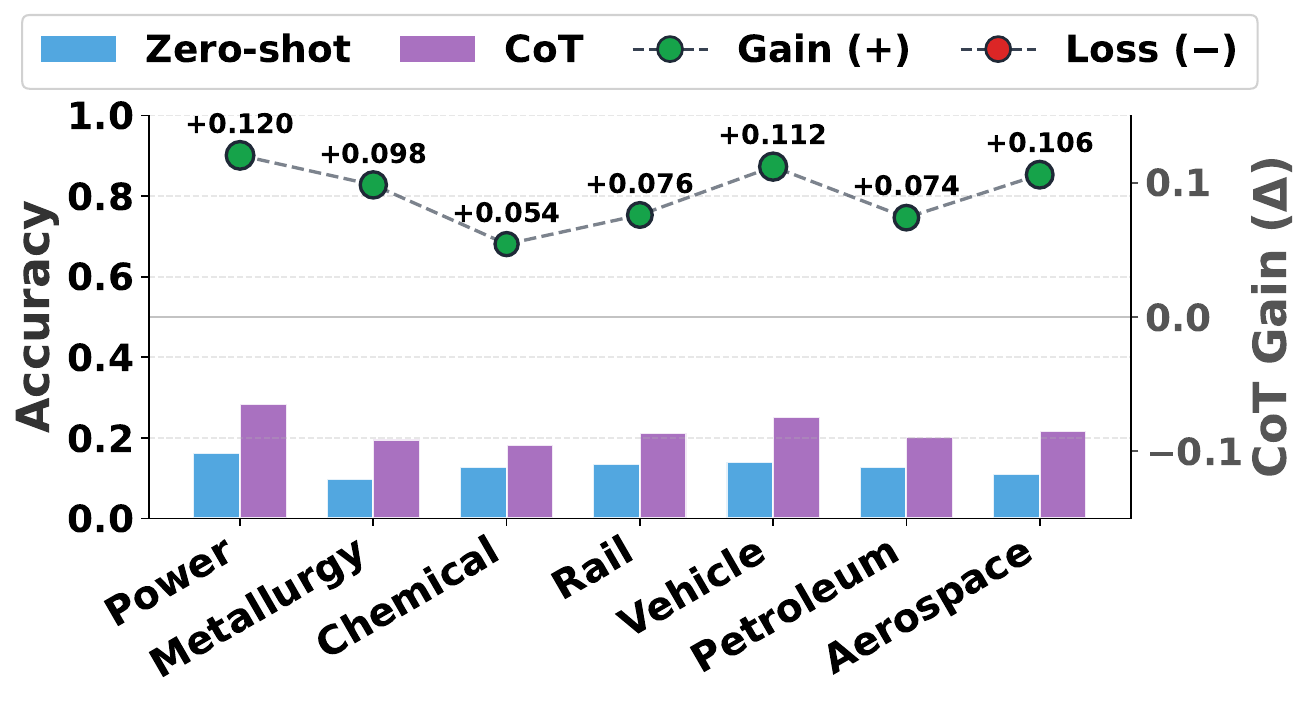}
        \caption{Qwen3-0.6B on L3 reasoning tasks.}
        \label{fig:cot_qwen_l3}
    \end{subfigure}
    \hfill
    \begin{subfigure}{0.45\linewidth}
        \centering
        \includegraphics[width=\linewidth]{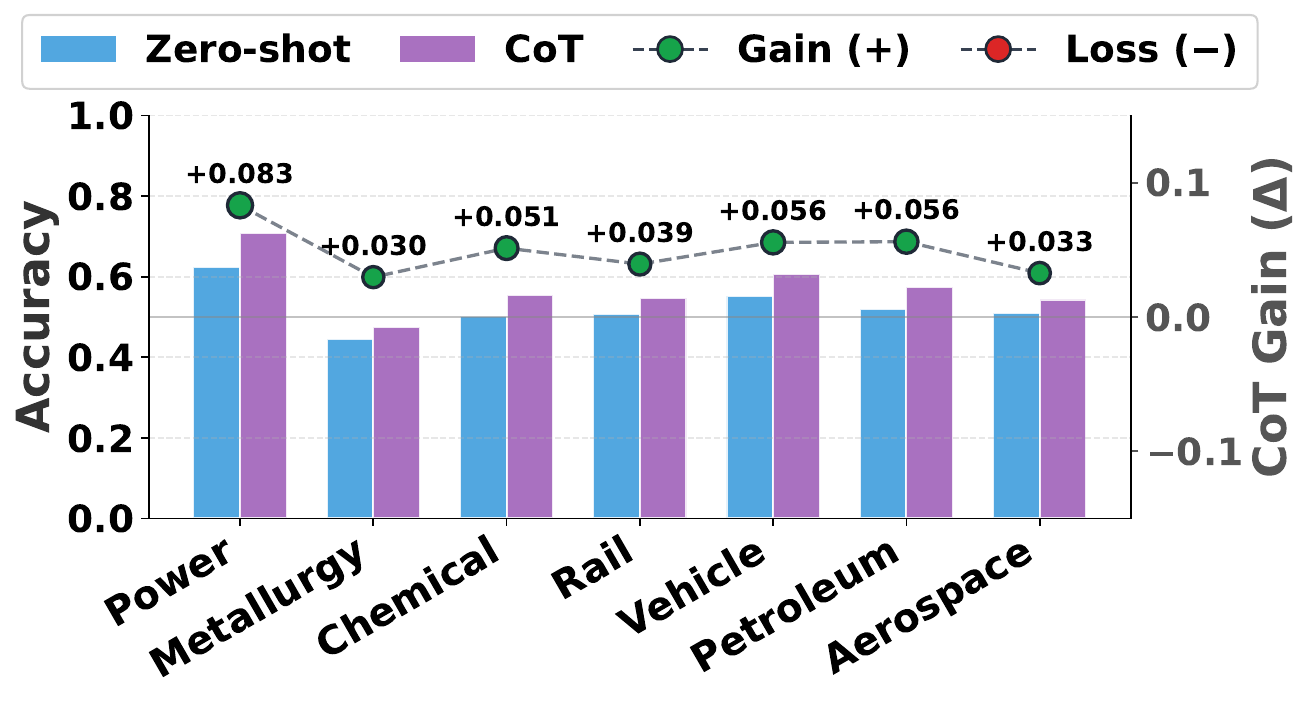}
        \caption{Qwen3-0.6B on L4 engineering application tasks.}
        \label{fig:cot_qwen_l4}
    \end{subfigure}

    \caption{\rev{Comparison of the zero-shot and CoT settings on L3 and L4 tasks for GPT-5 and Qwen3-0.6B.}}
    \label{fig:cot gain}
\end{figure}
\rev{Explicit reasoning benefits L3 more than L4: over the Chinese questions of the seven domains, the mean CoT gain is +0.028 on L3 and +0.006 on L4. On L3, the largest gains occur on Method Analysis (+0.034) and Process Flow Analysis (+0.033), but on L4 gains are small and vary in sign, from System Optimization (+0.015) to Condition-Impact Reasoning ($-$0.011).}

\rev{The effect of CoT varies by model. On the Chinese questions, 9 of the 11 models gain on L3 and 8 gain on L4. The largest losses occur for GLM-4.5 on L4 ($-$0.061), Gemini 2.5 Pro on L3 ($-$0.034), and GPT-4.1 on L4 ($-$0.033).}

\rev{Gains also decrease as zero-shot scores increase (Fig.~\ref{fig:inverse-scaling}; Pearson $r = -0.78$, $n = 11$). Qwen3-0.6B gains +0.092 on L3 and +0.049 on L4 (Fig.~\ref{fig:cot gain}), compared with +0.016 and +0.012 for GPT-5. The models with the largest gains are all from the Qwen series, so this trend also reflects model family.}

\rev{In Aerospace, the direction of the gain reverses with language (Fig.~\ref{fig:cot-heatmap}). On L3, CoT gains +0.029 on the Chinese questions but loses 0.029 on the English questions, where 8 of the 11 models score lower. On L4, the gain is +0.001 on the Chinese questions and +0.033 on the English questions, where all 11 models gain.}

\subsection{\rev{Pass@k Experiment}}\label{sec:passk}

\begin{figure}[H]
    \centering
    \begin{subfigure}{0.45\linewidth}
        \centering
        \includegraphics[width=\linewidth]{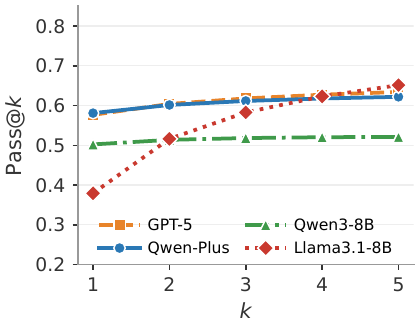}
        \caption{Pass@k-CN.}
        \label{fig:passk1}
    \end{subfigure}
    \hfill
    \begin{subfigure}{0.45\linewidth}
        \centering
        \includegraphics[width=\linewidth]{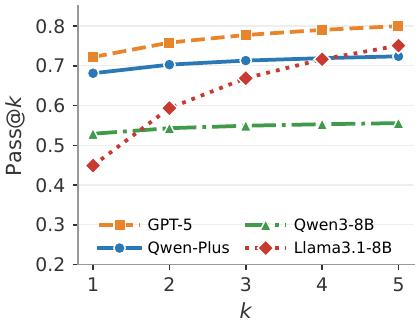}
        \caption{Pass@k-EN.}
        \label{fig:passk2}
    \end{subfigure}
    \caption{\rev{Pass@k curves ($k$=1 to 5) for GPT-5, Qwen-Plus, Qwen3-8B, and Llama-3.1-8B on L3 tasks in Chinese (CN) and English (EN).}}
    \label{fig:pass@k}
\end{figure}

\rev{We further conducted Pass@k ($k=1,\dots,5$) testing on L3 (Reasoning) tasks for GPT-5, Qwen-Plus, Qwen3-8B, and Llama-3.1-8B, with the sampling configuration and estimator described in Section~\ref{sec:setup}. As shown in Fig.~\ref{fig:pass@k}, on the Chinese questions Qwen-Plus and GPT-5 are close at Pass@1 (0.5808 vs. 0.5767), and GPT-5 is higher at $k=5$ (0.6344 vs. 0.6221). Qwen3-8B stays nearly the same (0.502 to 0.521), but Llama-3.1-8B rises from 0.379 to 0.652. The English questions show the same pattern: from $k=1$ to $k=5$, GPT-5 rises from 0.722 to 0.800, Qwen-Plus from 0.681 to 0.724, Qwen3-8B from 0.529 to 0.556, and Llama-3.1-8B from 0.449 to 0.751. The slope shows the consistency of the samples: all five samples were either correct or incorrect for 95.0\% of the items for Qwen3-8B, 91.8\% for Qwen-Plus, 88.6\% for GPT-5, and 43.6\% for Llama-3.1-8B.}

\subsection{Industrial Cognitive Capability Gap Analysis of LLMs}\label{sec:iccg}
\begin{figure}[!t]
    \centering
    \includegraphics[width=\linewidth]{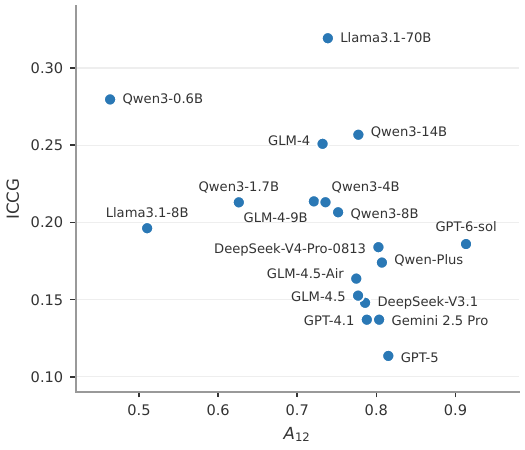}
    \caption{\rev{ICCG against foundational accuracy $A_{12}=\mathrm{Acc}(\mathrm{L1{+}L2})$ for the 18 evaluated models. Models with similar ICCG can differ widely in absolute performance.}}
    \label{fig:iccg}
\end{figure}
Table~\ref{tab:model-evaluation} reports ICCG together with absolute performance. GPT-5 has the smallest mean ICCG (0.113), followed by GPT-4.1 and Gemini 2.5 Pro (both 0.137), and Llama-3.1-70B has the largest (0.319). ICCG tends to decrease as $A_{12}$ increases (Fig.~\ref{fig:iccg}; Spearman $\rho=-0.67$, $n=18$), but a small gap does not imply high performance. GPT-6-sol has the highest $A_{12}$ (0.913) and $\mathrm{Acc}(\mathrm{L3{+}L4})$ (0.744), yet its ICCG (0.186) is larger than that of GPT-5. Llama-3.1-8B has a smaller ICCG (0.196) than Qwen3-14B (0.257), although its $A_{12}$ is much lower (0.510 vs. 0.777), and Qwen3-0.6B and Llama-3.1-70B have similar ICCG values (0.280 and 0.319) but very different $A_{12}$ (0.464 and 0.739).

\rev{The variation in ICCG across models is larger than the variation across domains: the model means range from 0.113 to 0.319, but the domain means range from 0.187 to 0.204.}

\begin{table*}[!t]
  \centering
  \footnotesize
  \renewcommand{\arraystretch}{1.3}
  \caption{\rev{ICCG of the models across domains, with absolute performance. ICCG $=1-r$ is computed per domain, and Aerospace includes its Chinese and English questions. Mean, $A_{12}$, and $\mathrm{Acc}(\mathrm{L3{+}L4})$ are averaged over the seven domains.}}
  \label{tab:model-evaluation}
  \resizebox{\linewidth}{!}{
  \begin{tabular}{c|ccccccc|c|cc}
  \toprule
  Models & Power & Metallurgy & Chemical & Rail Transit & Automotive & Petroleum & Aerospace & Mean & \rev{$A_{12}$} & \rev{$\mathrm{Acc}(\mathrm{L3{+}L4})$} \\
  \midrule
  \rev{DeepSeek-V3.1} & \rev{0.1854} & \rev{0.1163} & \rev{0.1478} & \rev{0.1404} & \rev{0.1275} & \rev{0.1524} & \rev{0.1655} & \rev{0.1479} & \rev{0.7858} & \rev{0.6688} \\
  \rev{DeepSeek-V4-Pro-0813} & \rev{0.1578} & \rev{0.2110} & \rev{0.1759} & \rev{0.1895} & \rev{0.1548} & \rev{0.1729} & \rev{0.2255} & \rev{0.1839} & \rev{0.8026} & \rev{0.6557} \\
  \rev{Gemini 2.5 Pro} & \rev{0.1477} & \rev{0.1375} & \rev{0.1449} & \rev{0.1217} & \rev{0.1343} & \rev{0.1569} & \rev{0.1157} & \rev{0.1370} & \rev{0.8035} & \rev{0.6934} \\
  \rev{GLM-4.5} & \rev{0.1605} & \rev{0.1335} & \rev{0.1534} & \rev{0.1521} & \rev{0.1331} & \rev{0.1676} & \rev{0.1675} & \rev{0.1525} & \rev{0.7769} & \rev{0.6581} \\
  \rev{GLM-4.5-Air} & \rev{0.1572} & \rev{0.1659} & \rev{0.1713} & \rev{0.1540} & \rev{0.1626} & \rev{0.1654} & \rev{0.1684} & \rev{0.1636} & \rev{0.7747} & \rev{0.6480} \\
  \rev{GLM-4} & \rev{0.1871} & \rev{0.2893} & \rev{0.2640} & \rev{0.2636} & \rev{0.2415} & \rev{0.2663} & \rev{0.2442} & \rev{0.2509} & \rev{0.7321} & \rev{0.5498} \\
  \rev{GLM-4-9B} & \rev{0.2073} & \rev{0.2274} & \rev{0.2170} & \rev{0.2091} & \rev{0.1926} & \rev{0.2094} & \rev{0.2325} & \rev{0.2136} & \rev{0.7210} & \rev{0.5674} \\
  \rev{GPT-4.1} & \rev{0.1425} & \rev{0.1337} & \rev{0.1640} & \rev{0.1141} & \rev{0.1409} & \rev{0.1518} & \rev{0.1117} & \rev{0.1369} & \rev{0.7880} & \rev{0.6799} \\
  \rev{GPT-5} & \rev{0.1232} & \rev{0.0963} & \rev{0.1198} & \rev{0.1059} & \rev{0.1189} & \rev{0.1357} & \rev{0.0945} & \rev{0.1135} & \rev{0.8151} & \rev{0.7224} \\
  \rev{GPT-6-sol} & \rev{0.1771} & \rev{0.1945} & \rev{0.1884} & \rev{0.1902} & \rev{0.1812} & \rev{0.1976} & \rev{0.1726} & \rev{0.1859} & \rev{0.9134} & \rev{0.7437} \\
  \rev{Llama-3.1-8B} & \rev{0.2048} & \rev{0.2422} & \rev{0.2123} & \rev{0.1778} & \rev{0.1791} & \rev{0.1827} & \rev{0.1747} & \rev{0.1962} & \rev{0.5103} & \rev{0.4108} \\
  \rev{Llama-3.1-70B} & \rev{0.2978} & \rev{0.3356} & \rev{0.3279} & \rev{0.3486} & \rev{0.3184} & \rev{0.3199} & \rev{0.2868} & \rev{0.3193} & \rev{0.7387} & \rev{0.5036} \\
  \rev{Qwen3-0.6B} & \rev{0.3997} & \rev{0.2162} & \rev{0.2776} & \rev{0.2397} & \rev{0.3108} & \rev{0.2947} & \rev{0.2187} & \rev{0.2796} & \rev{0.4635} & \rev{0.3290} \\
  \rev{Qwen3-1.7B} & \rev{0.1834} & \rev{0.2012} & \rev{0.1969} & \rev{0.1986} & \rev{0.1984} & \rev{0.1965} & \rev{0.3158} & \rev{0.2130} & \rev{0.6262} & \rev{0.4932} \\
  \rev{Qwen3-4B} & \rev{0.1970} & \rev{0.2507} & \rev{0.2183} & \rev{0.1885} & \rev{0.1901} & \rev{0.2152} & \rev{0.2315} & \rev{0.2130} & \rev{0.7357} & \rev{0.5797} \\
  \rev{Qwen3-8B} & \rev{0.1925} & \rev{0.2246} & \rev{0.2151} & \rev{0.1850} & \rev{0.1972} & \rev{0.2224} & \rev{0.2089} & \rev{0.2065} & \rev{0.7517} & \rev{0.5969} \\
  \rev{Qwen3-14B} & \rev{0.2352} & \rev{0.2798} & \rev{0.2665} & \rev{0.2542} & \rev{0.2310} & \rev{0.2654} & \rev{0.2652} & \rev{0.2568} & \rev{0.7773} & \rev{0.5783} \\
  \rev{Qwen-Plus} & \rev{0.1720} & \rev{0.1896} & \rev{0.1973} & \rev{0.1646} & \rev{0.1595} & \rev{0.1934} & \rev{0.1413} & \rev{0.1739} & \rev{0.8070} & \rev{0.6669} \\
  \midrule
  Mean & \rev{0.1960} & \rev{0.2025} & \rev{0.2032} & \rev{0.1888} & \rev{0.1873} & \rev{0.2037} & \rev{0.1967} & \rev{0.1969} & \rev{0.7402} & \rev{0.5970} \\
  \bottomrule
  \end{tabular}
  }
\end{table*}

\rev{How the models are ordered depends on which gap is measured. Without L4, the gap widens for 17 of the 18 models and the mean ICCG rises from 0.197 to 0.325. Because this variant rests on L3 alone, it captures a narrower quantity and orders the models differently from the revised ICCG (Spearman $\rho=0.692$). Llama-3.1-8B, for instance, has the smallest L3-only gap even though its foundational accuracy is only $A_{12}=0.510$, a reminder that a relative gap means little without the absolute scores beside it. Agreement is higher with the absolute difference $\Delta=A_{12}-\mathrm{Acc}(\mathrm{L3{+}L4})$ ($\rho=0.779$), under which GPT-5 again has the smallest gap, and higher still when GPT-6-sol or Qwen3.7-Plus scores L4 ($\rho=0.880$ and 0.897). For this reason we base no conclusion on small differences in rank. Appendix~\ref{sec:iccg_variants} gives the details.}

\section{Discussion}

\rev{We read these results as a profile of industrial knowledge, and not as a decline along levels of increasing difficulty. Format and scoring change both the size and the direction of the level-wise differences, so the four levels describe types of cognitive demand. Within this profile, the specific knowledge also matters. Data Extraction is nominally a memory task, but it is more error-prone than many higher-level tasks. Many of these weaknesses are common across models. Newer and larger models achieve higher scores, but the most difficult task types remain the same for all 18 models. Models with similar overall scores can behave differently. On L3, the GPT, GLM, and Qwen families tend to select too many options, but the DeepSeek family more often misses correct ones. CoT also affects models differently, and its effect depends on the model, the level, and the language. ICCG adds to this profile something that absolute scores do not show, because it separates models that have similar foundational accuracy but different higher-order performance. It also varies more across models than across domains, and its model ranking largely stays the same when the two additional judges re-score L4.}

\rev{IndustrialKnowEval measures industrial knowledge and engineering question answering. All results are obtained in a closed-book setting and characterize standalone models rather than systems augmented with retrieval or other tools. The benchmark does not test whether a model can use manuals, retrieval systems, and other tools within a diagnostic workflow. These capabilities require evaluation protocols based on real engineering cases, which we leave to future work. One possible route is to work with enterprise specialists to convert desensitized maintenance work orders, fault reports, and alarm sequences into multi-source, multi-turn diagnostic cases, with expert-annotated reference solutions and an assessment of whether the model requests missing information.}

\section{Conclusion}
\rev{IndustrialKnowEval brings together 43,282 questions from seven industrial domains, and we built it through expert-guided task design, source traceability, and multi-stage quality control. It evaluates industrial knowledge and engineering question answering, the knowledge foundation that industrial LLM applications build on. Its core is the L1--L4 framework, whose 21 task types run from foundational memory to engineering application. When we evaluated 18 LLMs on it, mainstream models did better on L1 and L2, but L3 reasoning was the lowest-scoring level. Even the strongest models had trouble with specific numerical values and with conceptual and process-related knowledge. Under the primary judge, L4 engineering application scored higher than L3. We further examined its scoring with two additional judges and three experts. The proposed ICCG metric quantifies the gap between foundational and higher-order performance, and together with the error analysis exposes specific knowledge blind spots across models and domains.}

Looking forward, IndustrialKnowEval will evolve to integrate multimodal data fusion and \rev{real engineering cases with multi-turn reasoning}, advancing evaluation toward system-level engineering contexts. We \rev{believe} that a robust, multi-tiered evaluation system \rev{can provide clear guidance} for optimizing industrial LLMs \rev{and a solid foundation for bringing them into} real-world scenarios.

\bibliographystyle{ieeetr}
\bibliography{reference}

\clearpage
\begin{IEEEbiography}[{\includegraphics[width=1in,height=2.5in,clip,keepaspectratio]{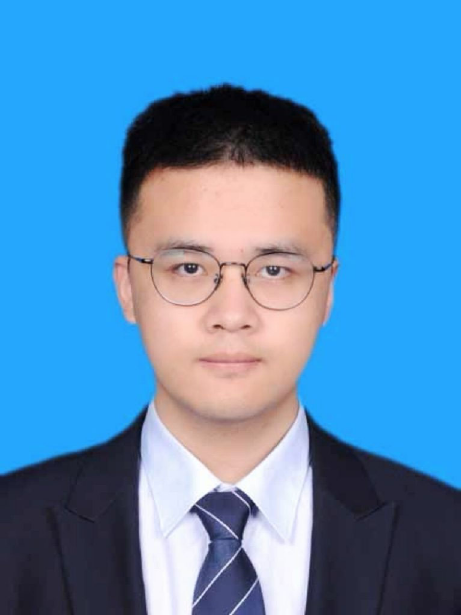}}]{Haiteng Wang} (Member, IEEE) received the Ph.D. degree in pattern recognition and intelligent systems from the School of Automation Science and Electrical Engineering, Beihang University, Beijing, China, in 2026. He is currently an Assistant Professor with Beihang University. His current research interests include Time Series Prediction, Industrial Foundation Model, Generative Models.
\end{IEEEbiography}

\begin{IEEEbiography}[{\includegraphics[width=1in,height=2.5in,clip,keepaspectratio]{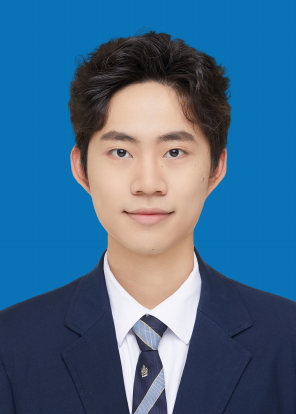}}]{Haoze Li} received the B.S. degree in Aircraft Control and Information Engineering from the School of Aeronautic Science and Engineering, Beihang University, Beijing, China, in 2025. He is currently pursuing the M.S. degree under the supervision of Prof. Lei Ren. He is also a student at
the National College for Excellent Engineers, Beihang University. His research interests include industrial artificial intelligence, large language models, and intelligent agents.
\end{IEEEbiography}

\begin{IEEEbiography}[{\includegraphics[width=1in,height=2.5in,clip,keepaspectratio]{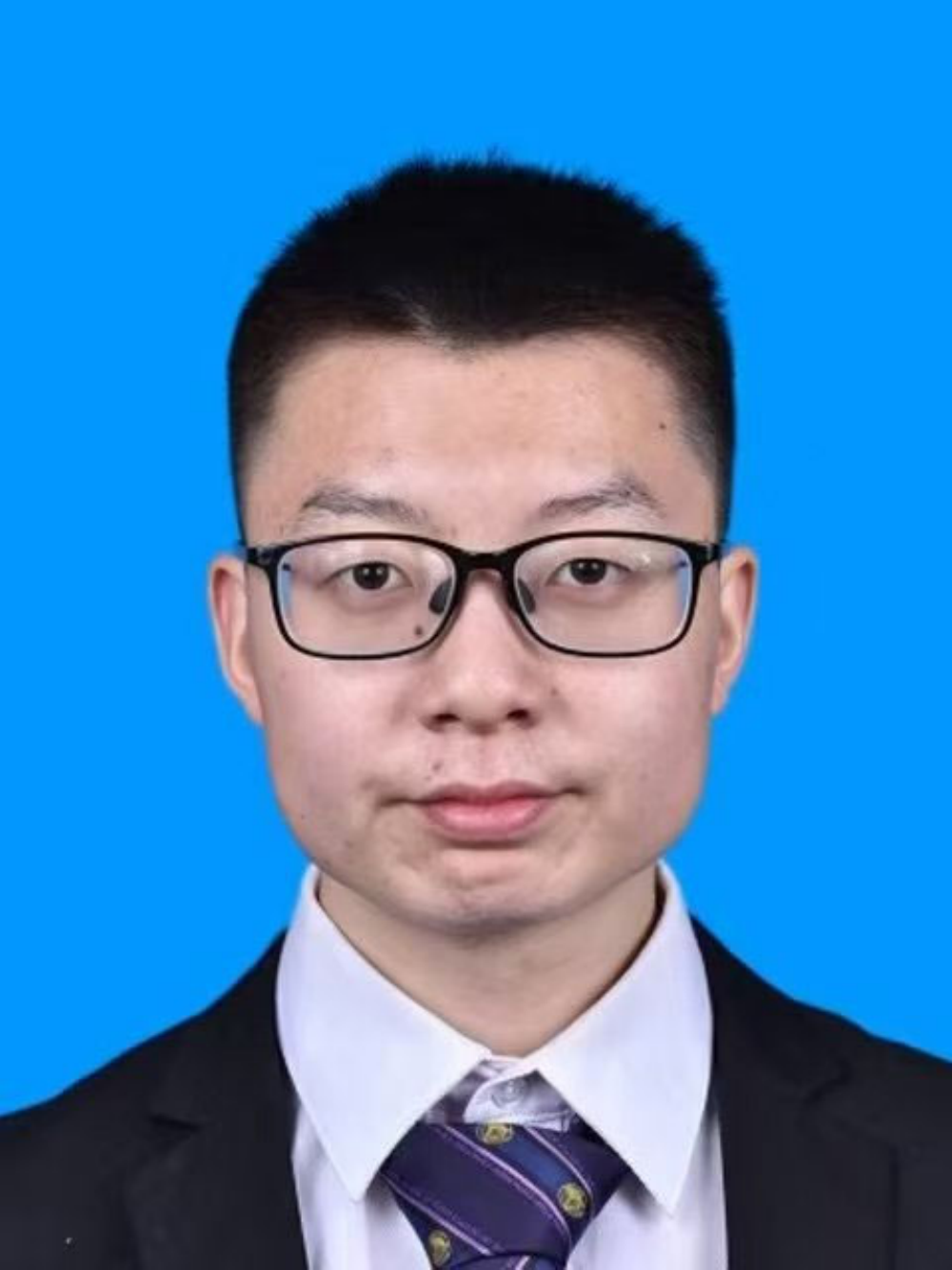}}]{Jingwei Guo}(Student Member, IEEE) received the B.S. degree from the School of Automation, Beijing Institute of Technology, Beijing, China, in 2023, where he is currently pursuing the Ph.D. degree, supervised by Prof. Lei Ren. He is an Engineer with Zhongguancun Laboratory, Beijing. His research interests include industrial artificial intelligence, time series analysis and the IIoT cyber security.
\end{IEEEbiography}

\begin{IEEEbiography}[{\includegraphics[width=1in,height=2.5in,clip,keepaspectratio]{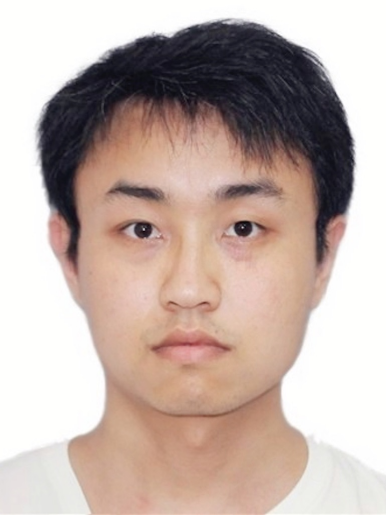}}]{Jingbo Zhou} is the Agent Frontier Team lead in Baidu’s ERNIE foundation model department and head of the Business Intelligence Lab at Baidu Research. He holds a Ph.D. in computer science from the National University of Singapore (2014) and a B.E. from Shandong University (2009). His recent research focuses on advancing LLM agents for real-world tasks through new model architectures, post-training methods such as agentic reinforcement learning, and harness design. He has published over 100 papers in venues including KDD, SIGMOD, AAAI, Nature Machine Intelligence, and Nature Sustainability. He regularly serves as an area chair or senior program committee member at conferences including KDD, AAAI, IJCAI, and WWW. His work has contributed to ERNIE, Baidu Maps, and Baidu Search Ads.
\end{IEEEbiography}

\begin{IEEEbiography}[{\includegraphics[width=1in,height=2.5in,clip,keepaspectratio]{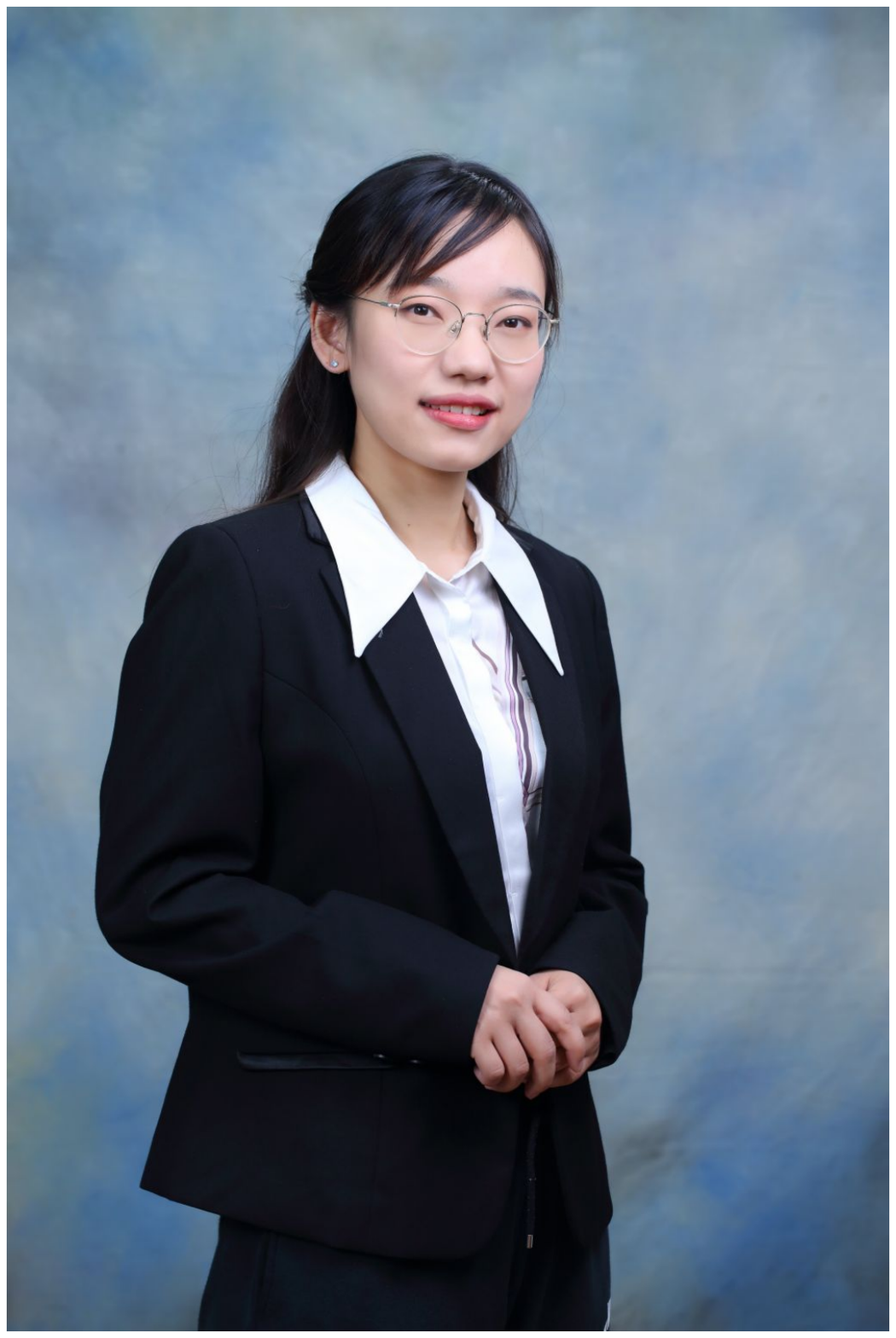}}]{Si Liu} (Member, IEEE) is currently a full professor and vice dean of the School of Artificial Intelligence, Beihang University. She is a National Leading Talent and is the recipient of the Young Scientists Fund (B Class). Her research interests include cross-modal multimedia intelligent analysis and embodied intelligence. She has published more than 100 cutting-edge papers and has been cited over 20,000 times on Google Scholar. She has won the Best Paper Awards of ACM MM 2021 and 2013, the Best Video Award of IJCAI 2021, and the Best Demo Award of ACM MM 2012. She is currently the associate editor of IEEE TPAMI, IEEE TMM, IEEE TCSVT, and CVIU. She has served as the area chair of NeurIPS, ICML, ICCV, CVPR, and other top conferences many times.
\end{IEEEbiography}

\begin{IEEEbiography}[{\includegraphics[width=1in,height=2.5in,clip,keepaspectratio]{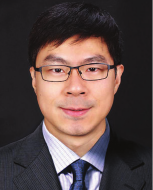}}]{Lei Ren} (Senior Member, IEEE) received the Ph.D. degree in computer science from the Institute of Software, Chinese Academy of Sciences, Beijing, China, in 2009. 
	
He is currently a Professor with the School of Automation Science and Electrical Engineering, Beihang University, Beijing, China, also with the Hangzhou International Innovation Institute, Beihang University, Hangzhou, China, and also with the State Key Laboratory of Intelligent Manufacturing System Technology, Beijing, China. His research interests include neural networks and deep learning, time series analysis, and industrial AI applications. Dr. Ren serves as an Associate Editor for the IEEE Transactions on Neural Networks and Learning Systems, IEEE/ASME Transactions on Mechatronics and other international journals.	
\end{IEEEbiography}
\vskip 0pt plus 1filll
\clearpage

\appendices
\section{Dataset Overview}
\begin{table}[!h]
  \centering
  \footnotesize
  \renewcommand{\arraystretch}{1.3}
  \caption{Overview of the IndustrialKnowEval Dataset Across Seven Industrial Domains and Four Cognitive Levels. SCQ means Single-Choice Question, MCQ means Multiple-Choice Question, and QA means Open-ended Question.}
  \label{tab:dataoverview}
  \resizebox{\linewidth}{!}{
  \begin{tabular}{c|c|ccccccc|c}
  \toprule
  Level & Task Type & Aerospace & Power & Metallurgy & Chemical & Automotive & Rail Transit & Petroleum & Total\\
  \midrule

  \multirow{5}{*}{\makecell{L1\\SCQ}}
  & Method Recognition & 406 & 636 & 409 & 448 & 419 & 240 & 499  & \multirow{5}{*}{10945}\\
  & Data Extraction & 271 & 137 & 751 & 579 & 209 & 203 & 553 \\
  & Concept Recall & 501 & 598 & 199 & 274 & 253 & 127 & 301  \\
  & Entity Recognition & 60 & 33 & 87 & 112 & 55 & 97 & 97 \\
  & Category Identification & 702 & 350 & 285 & 358 & 267 & 33 & 396 \\

  \midrule

  \multirow{5}{*}{\makecell{L2\\SCQ}}
  & Principle Explanation & 815 & 756 & 959 & 857 & 466 & 368 & 936 & \multirow{5}{*}{10919}\\
  & Function Description & 736 & 897 & 274 & 460 & 516 & 212 & 557 \\
  & Structural Feature Description & 91 & 21 & 77 & 50 & 75 & 49 & 64 \\
  & Process Description     & 192 & 30 & 299 & 314 & 118 & 43 & 194 \\
  & Formula Comprehension     & 101 & 28 & 131 & 84 & 29 & 36 & 84 \\

  \midrule

  \multirow{6}{*}{\makecell{L3\\MCQ}}
  & Influence Factor Analysis & 757 & 629 & 842 & 853 & 506 & 437 & 991  & \multirow{6}{*}{10930}\\
  & Method Analysis     & 561 & 345 & 192 & 210 & 194 & 159 & 219 \\
  & Performance Comparison     & 510 & 721 & 479 & 514 & 438 & 89 & 489  \\
  & Process Flow Analysis & 34 & 19 & 145 & 125 & 45 & 12 & 82\\
  & Phenomenon Interpretation & 40 & 6 & 31 & 26 & 7 & 11 & 23\\
  & Structure Function Analysis & 34 & 34 & 31 & 51 & 15 & 0 & 24 \\

  \midrule

  \multirow{5}{*}{\makecell{L4\\QA}}
  & Root Cause Reasoning     & 873 & 876 & 1205 & 1189 & 665 & 386 & 1227  & \multirow{5}{*}{10488}\\
  & Condition-Impact Reasoning & 387 & 303 & 213 & 207 & 152 & 94 & 224 \\
  & System Optimization & 402 & 216 & 149 & 182 & 224 & 88 & 174 \\
  & Fault Diagnosis     & 87 & 158 & 21 & 40 & 52 & 85 & 40 \\
  & Consequence Prediction & 170 & 79 & 91 & 68 & 50 & 38 & 73 \\

  \midrule

  Total & -- & 7730 & 6872 & 6870 & 7001 & 4755 & 2807 & 7247 & 43282\\
  \bottomrule
  \end{tabular}
  }
\end{table}

\begin{table}[!h]
  \centering
  \footnotesize
  \renewcommand{\arraystretch}{1.3}
  \caption{\rev{Source Documents and Questions of IndustrialKnowEval by Domain. Q/Doc is the mean number of retained questions per source document. Top-10 is the share of a domain's questions derived from its ten most frequently used documents.}}
  \label{tab:sourcestats}
  \resizebox{\linewidth}{!}{
  \begin{tabular}{l|cc|cc|c|cccc|c}
  \toprule
  \multirow{2}{*}{Domain} & \multicolumn{2}{c|}{Documents} & \multicolumn{2}{c|}{Questions} & \multirow{2}{*}{Q/Doc} & \multicolumn{4}{c|}{Questions by Level} & \multirow{2}{*}{Top-10 (\%)} \\
  & Count & Share (\%) & Count & Share (\%) & & L1 & L2 & L3 & L4 & \\
  \midrule
  Aerospace & 2,000 & 16.5 & 7,730 & 17.9 & 3.87 & 1,940 & 1,935 & 1,936 & 1,919 & 0.58 \\
  Power & 2,025 & 16.7 & 6,872 & 15.9 & 3.39 & 1,754 & 1,732 & 1,754 & 1,632 & 0.58 \\
  Metallurgy & 2,008 & 16.5 & 6,870 & 15.9 & 3.42 & 1,731 & 1,740 & 1,720 & 1,679 & 0.60 \\
  Chemical & 2,012 & 16.6 & 7,001 & 16.2 & 3.48 & 1,771 & 1,765 & 1,779 & 1,686 & 0.59 \\
  Automotive & 1,327 & 10.9 & 4,755 & 11.0 & 3.58 & 1,203 & 1,204 & 1,205 & 1,143 & 0.84 \\
  Rail Transit & 765 & 6.3 & 2,807 & 6.5 & 3.67 & 700 & 708 & 708 & 691 & 1.43 \\
  Petroleum & 2,014 & 16.6 & 7,247 & 16.7 & 3.60 & 1,846 & 1,835 & 1,828 & 1,738 & 0.57 \\
  \midrule
  Total & 12,151 & 100.0 & 43,282 & 100.0 & 3.56 & 10,945 & 10,919 & 10,930 & 10,488 & -- \\
  \bottomrule
  \end{tabular}
  }
\end{table}

\section{Model Details}
\label{modeldetails}
\rev{Table~\ref{tab:models} lists the 18 evaluated models with their versions, access interfaces, and reasoning settings in the zero-shot evaluation. The zero-shot and CoT evaluations use temperature~0. In the CoT setting, the thinking mode of the Qwen series, DeepSeek-V3.1, and GLM-4.5 is enabled, and the reasoning effort of GPT-5 and Gemini~2.5~Pro is raised from the lowest level to the default. The Pass@k experiment samples at temperature~1, with top-$p$ and the random seed left at their API defaults. The primary L4 judge is Qwen-Plus(qwen-plus-2025-09-11), run with thinking disabled at the provider's default temperature.}

\begin{table}[!h]
\centering
\renewcommand{\arraystretch}{1.3}
\caption{\rev{The 18 LLMs evaluated in our experiments. Reasoning is the setting in the zero-shot evaluation; ``--'' means that the model has no adjustable reasoning setting.}}
\label{tab:models}
\resizebox{\linewidth}{!}{
\begin{tabular}{lllcllc}
\toprule
\textbf{ID} & \textbf{Model} & \textbf{Creator} & \textbf{\#Parameters} & \textbf{API Version} & \textbf{Access} & \textbf{Reasoning} \\
\midrule
M1  & GPT-5 & OpenAI & \textit{undisclosed} & openai/gpt-5 & OpenRouter & Minimal effort \\
M2  & GPT-4.1 & OpenAI & \textit{undisclosed} & openai/gpt-4.1 & OpenRouter & -- \\
M3  & Gemini~2.5~Pro & Google & \textit{undisclosed} & google/gemini-2.5-pro & OpenRouter & Lowest effort \\
M4  & DeepSeek-V3.1 & DeepSeek AI & 671B & deepseek-chat & DeepSeek API & Thinking off \\
M5  & GLM-4.5 & ZhipuAI & \textit{undisclosed} & glm-4.5 & Zhipu BigModel & Thinking off \\
M6  & GLM-4.5-Air & ZhipuAI & \textit{undisclosed} & glm-4.5-air & Zhipu BigModel & -- \\
M7  & GLM-4 & ZhipuAI & \textit{undisclosed} & glm-4 & Zhipu BigModel & -- \\
M8  & Qwen-Plus & Alibaba Cloud & \textit{undisclosed} & qwen-plus-2025-09-11 & Alibaba Cloud & Thinking off \\
\midrule
M9  & Qwen3-0.6B & Alibaba Cloud & 0.6B & qwen3-0.6b & Alibaba Cloud & Thinking off \\
M10 & Qwen3-1.7B & Alibaba Cloud & 1.7B & qwen3-1.7b & Alibaba Cloud & Thinking off \\
M11 & Qwen3-4B & Alibaba Cloud & 4B & qwen3-4b & Alibaba Cloud & Thinking off \\
M12 & Qwen3-8B & Alibaba Cloud & 8B & qwen3-8b & Alibaba Cloud & Thinking off \\
M13 & Qwen3-14B & Alibaba Cloud & 14B & qwen3-14b & Alibaba Cloud & Thinking off \\
M14 & Llama-3.1-8B & Meta & 8B & meta-llama/llama-3.1-8b-instruct & OpenRouter & -- \\
M15 & Llama-3.1-70B & Meta & 70B & meta-llama/llama-3.1-70b-instruct & OpenRouter & -- \\
M16 & GLM-4-9B & ZhipuAI & 9B & glm-4-9b & Zhipu BigModel & -- \\
\midrule
\rev{M17} & \rev{DeepSeek-V4-Pro-0813} & \rev{DeepSeek AI} & \rev{\textit{undisclosed}} & \rev{deepseek-v4-pro} & \rev{DeepSeek API} & \rev{Thinking off} \\
\rev{M18} & \rev{GPT-6-sol} & \rev{OpenAI} & \rev{\textit{undisclosed}} & \rev{gpt-6-sol} & \rev{OpenRouter} & \rev{Thinking off} \\

\bottomrule
\end{tabular}
}
\end{table}
\vfill\newpage

\section{Prompt}
\label{prompt}
This appendix presents the prompt templates used for question generation, model evaluation, L4 grading, and task-type labeling. We show the English versions; Chinese-language evaluations used semantically equivalent prompts translated by bilingual domain experts. The target roles and difficulty descriptions in the generation templates are instructions to the generator. The four levels themselves are defined by the type of cognitive demand (Section~\ref{sec:cognitive}). All prompts are also available in our public code repository.

\begin{prompt}{Prompt Templates of Different Types}
L1 -  Memory level

Target role: Apprentice or hobbyist.

Focus: Recall and identification of terms, values, or categories.

Typical verbs: "What is...", "How much is...", "Which includes..."

Difficulty: Very simple factual recall.

Only one correct answer. Ensure A/B/C/D are roughly evenly distributed across questions.
\\
\\
L2 - comprehension level

Target role: Junior engineer or professional undergraduate.

Focus: Comprehension of steps, formulas, or methods.

Typical verbs: "What represents...", "What is the sequence...", "Which method is used..."

Difficulty: Moderate.

Only one correct answer. Maintain balanced answer distribution (A-D).
\\
\\
L3 - Reasoning level

Target role: Intermediate engineer or researcher.

Focus: Analytical reasoning, comparison, cause-effect analysis.

Typical verbs: "What are the advantages...", "What is the rationale...", "Why does..."

Difficulty: Medium.

The correct answer must include at least two options, separated by commas (e.g., "A,C").
\\
\\
L4 - Engineering Application level

Target role: Senior engineer or expert.

Focus: Deep reasoning - root cause analysis, multi-step inference, or system-level coupling.

Typical verbs: ``What is the root cause...'', ``What leads step by step to...'', ``Why does the fault occur...''

Difficulty: Upper-intermediate.

The answer must synthesize multiple evidence fragments logically and coherently.

\end{prompt}

\begin{prompt}{Prompt Templates for Task Construction}
\textbf{ROLE}

You are a professional academic exam designer specializing in the aerospace field, skilled at generating high-quality questions based on research content. Based on the following professional knowledge, please create one L1 choice question/ one L2 choice question/ one L3 choice question/ one L4 open-ended question.
\\
\\
\textbf{PROFESSIONAL KNOWLEDGE}

[{text}]
\\
\\
\textbf{REQUIREMENTS}
\\
\\
\textbf{General Rules for All Questions:}

1. Each question must be clear, specific, and self-contained.

2. Questions should focus on key concepts, principles, mechanisms, methods, or conclusions, rather than peripheral or context-specific details.

3. Avoid referential or deictic expressions such as "this paper", "the article", "it", or "as mentioned above."

4. When referring to experiments or simulations, name them explicitly.

5. Each of the four questions must test different knowledge fragments from the provided text. Avoid reusing the same sentence, figure, or parameter across multiple questions.
\\
\\
\textbf{Additional Rules for Choice Questions}

1. Provide exactly four options: A, B, C, D.

2. Ensure all options share the same grammatical form.

3. Option length balance: keep options roughly the same length to prevent answer exposure by length cues.

4. Incorrect options should be plausible distractors, based on typical misconceptions or near-correct variants - not random or irrelevant statements.

5. Option D must not be an inclusive option such as "All of the above" or "Both A and B."

6. Each multiple-choice question must include an explanation.
\\
\\
\textbf{Question Type Specifications}

\{Prompt Templates of Different Types\}
\\
\\
\textbf{Final Quality Check}

Before submission, ensure that:

1. Each question is self-contained, with no context-dependent expressions.

2. Multiple-choice options are grammatically and visually balanced.

3. The explanation aligns exactly with the correct answer.

4. The knowledge reference directly supports the answer and is quoted verbatim.

5. If any issue is detected, regenerate the question until all requirements are met.

6. \textbf{Output only the final valid version.}
\\
\\
\textbf{OUTPUT FORMAT}

For each question, do not include any Markdown symbols such as "\#", "*", or "-".
Separate questions with one blank line.
For single-choice or multiple-choice questions, use the following format; For open-ended questions, keep the same format, but fill "options" and "explanation" with "None":
"difficulty": "Level",

"question": "Question text",

"options": "A. ...; B. ...; C. ...; D. ...",

"correct\_answer": "Correct option(s)",

"explanation":"Explanation text",
\\
\\
\textbf{EXAMPLE}

Some examples.

\end{prompt}

\begin{prompt}{Prompts for Evaluating L1 and L2 Tasks}
You are an expert in the industrial manufacturing domain. For the following choice question, analyze it based on your professional knowledge and provide the correct answer.

Strict Output Requirements:

1. Respond with \textbf{only the letter} of the correct option.

2. Do not include any explanation, analysis, or any other text or characters.
\\
\\
Question:
\\
\\
Options:
\end{prompt}

\begin{prompt}{Prompts for Evaluating L3 Tasks}
You are an expert in the industrial manufacturing domain. For the following multiple-choice question, analyze it based on your professional knowledge and provide the correct answers.

Strict Output Requirements:

1. Respond with \textbf{only the letters} of the correct options, separated by commas.

2. Do not include any explanation, analysis, or any other text or characters.
\\
\\
Question:
\\
\\
Options:
\end{prompt}

\begin{prompt}{Prompts for Evaluating L4 Tasks}
You are an expert in the industrial manufacturing domain. For the following open-ended question, analyze it based on your professional knowledge and provide a \textbf{concise and accurate} answer.

Strict Output Requirements:

1. Your answer must be direct and to the point.

2. Do not restate or repeat the question in your response.
\\
\\
Question:
\end{prompt}

\begin{prompt}{CoT Prompts for Evaluating L3 Tasks}
You are a logical and rigorous expert in the industrial manufacturing domain, adept at thinking through problems step-by-step to provide correct answers.

For the following multi-choice question, \textbf{think step-by-step} to analyze it based on your professional knowledge and finally provide the correct answers.

Strict Output Requirements:

1. Respond with \textbf{only the letters} of the correct options, separated by commas.

2. Do not include any thinking process, explanation, analysis, or any other text or characters in your final response.
\\
\\
Question:
\\
\\
Options:
\end{prompt}

\begin{prompt}{CoT Prompts for Evaluating L4 Tasks}
You are a logical and rigorous expert in the industrial manufacturing domain, adept at thinking through problems step-by-step to provide correct answers.

For the following open-ended question, \textbf{think step-by-step} to analyze it based on your professional knowledge and provide a \textbf{concise and accurate} answer.

Strict Output Requirements:

1. Your answer must be direct and to the point.

2. Do not restate or repeat the question in your response.
\\
\\
Question:
\end{prompt}

\begin{prompt}{Prompt for Evaluating Model's Answers of L4 Tasks}
You are an expert evaluator in the aerospace domain. Your task is to grade the [Model Answer] based on the [Question] and the [Reference Answer]. The score must range from 1 to 10, where 1 indicates a completely irrelevant or incorrect response, and 10 indicates a highly accurate and complete response.
\\
\\
The [Reference Answer] represents the ground-truth; all evaluation criteria must rely solely on the [Reference Answer], without introducing your own assumptions.
\\
\\
\texttt{[Question]}: \{question\}

\texttt{[Reference Answer]}: \{correct\_answer\}

\texttt{[Model Answer]}: \{model\_answer\}
\\
\\
Please assess the [Model Answer] according to the following criteria, in order:

1. Accuracy: Check whether the key facts, data, and concepts match the [Reference Answer]. If the [Model Answer] contains factual errors, the score must not exceed 5.

2. Completeness: If the [Reference Answer] includes multiple essential points, verify whether the [Model Answer] covers all of them.

3. Relevance: Ensure the [Model Answer] stays focused on the Question and does not drift off-topic.

4. Conciseness: The [Model Answer] should be succinct, without unnecessary or extraneous content.
\\
\\
Please output the result in the following format:

Score: an integer from 1 to 10

Reason: a brief justification
\end{prompt}
\rev{
The two additional judges, GPT-6-sol and Qwen3.7-Plus, use the same prompt, 1--10 scale, and normalization as the primary judge.
}
\vfill\newpage
\begin{prompt}{Task Type Definitions}
\begin{CJK}{UTF8}{gbsn}

L1

概念识记(Concept Recall): The question requires recalling definitions of terms, symbols, or other basic knowledge.

数值抽取(Data Extraction): The question requires extracting numerical values, parameters, or measurements; the answer is a specific number.

分类归属(Category Identification): The question requires determining or selecting the category, type, or classification of an item.

方法识别(Method Recognition): The question requires identifying a described technique, method, source, or characteristic.

工业实体识别(Entity Recognition): The question requires identifying the name or type of an entity, device, material, tool, or structure.
\\
\\
L2

公式理解(Formula Comprehension): The question requires explaining the meaning of a formula, the definition of variables, or the applicable scenarios of the formula.

工艺描述(Process Description): The question requires understanding or describing the workflow or steps of a process or engineering procedure.

原理解释(Principle Explanation): The question requires explaining the principle, phenomenon, or working mechanism of a technique or system.

功能描述(Function Description): The question requires understanding or describing the function, purpose, or operation of a device or instrument.

结构特征描述(Structural Feature Description): The question requires describing the structural features or components of an object, device, or system.
\\
\\
L3

性能比较(Performance Comparison): The question requires comparing the performance differences of methods, devices, parameters, or materials.

方法分析(Method Analysis): The question requires analyzing the applicability, advantages, disadvantages, or effectiveness of a method, technique, or strategy.

工艺流程分析(Process Flow Analysis): The question requires analyzing the steps of a process and the relationships between them.

实验现象分析(Phenomenon Interpretation): The question requires analyzing experimental or computational phenomena or observations and explaining their causes or significance.

影响因素分析(Influence Factor Analysis): The question requires analyzing the main factors affecting a phenomenon, result, or process and their mechanisms.

结构功能分析(Structure Function Analysis): The question requires analyzing how the structure of a device, material, or system relates to its function.
\\
\\
L4

影响预测(Consequence Prediction): The question requires forward reasoning to predict the consequences or effects of an operation, decision, or change.

系统优化设计(System Optimization): The question requires optimizing a system to improve its performance or efficiency.

故障排查(Fault Diagnosis): The question requires analyzing the cause of a system or device failure and proposing diagnostic steps.

条件影响推理(Condition-Impact Reasoning): The question requires analyzing how changes in certain conditions affect a system, process, or outcome, and explaining the mechanism.

根因推理(Root Cause Reasoning): The question requires analyzing the fundamental cause of a complex phenomenon or problem, tracing step by step from results or observations.
\end{CJK}
\end{prompt}
\begin{prompt}{Prompt for Labeling}
You are an expert in question classification. Based on the content of the question, determine which of the following types it belongs to, and output only the type name.

If the question does not fit any of the provided types, create a new type yourself and output its name.
\\
\\
Difficulty Level: 

\{difficulty\}

Candidate Types: 

\{type\_list\}

Question:

\{question\_text\}

Additional Information:

\{extra\_info\}
\\
\\
Output Requirements:

Output only the \textbf{type name} (it must be one of the provided types or a newly created type).

Do not include any extra text, punctuation, or explanations.
\end{prompt}

\clearpage
\section{Quality Control}
\label{qualitycontrol}

\subsection{Criteria for Identifying Context-Dependent Factual Reasoning (CDFR) Questions}

To ensure that the benchmark evaluates models' intrinsic industrial knowledge and reasoning capabilities rather than context-dependent information retrieval, we established a standardized filtering process to identify and exclude Context-Dependent Factual Reasoning (CDFR) questions. A question is classified as a CDFR item if it satisfies any one of the following three criteria:

\vspace{0.5em}
\noindent\rev{\textbf{Criterion 1: Specific Data Dependency.} The correct answer depends on a specific numerical value, parameter, or measurement. The value is not a universally established constant or a widely recognized engineering standard, and it can only be found in the source document or experimental context behind the question. \textit{Annotation indicator:} Annotators mark ``yes'' when the question cannot be answered correctly without the specific numerical data in the source text.}

\vspace{0.5em}
\noindent\rev{\textbf{Criterion 2: Specific Named Entity Dependency.} The correct answer depends on identifying or recalling a named entity, such as a proprietary system name, a project-specific device designation, a unique experimental apparatus, or a named individual. The entity is not well known or publicly recognized in the relevant industrial domain, and only the source document mentions it. \textit{Annotation indicator:} Annotators mark ``yes'' when the question cannot be answered without prior exposure to the document in which the named entity appears.}

\vspace{0.5em}
\noindent\rev{\textbf{Criterion 3: Implicit Context Dependency.} The question presupposes a specific experimental setup, a particular case study, or a specialized local context that only the source document describes. It may use referential expressions such as ``the proposed method,'' ``the described system,'' or ``the above experiment,'' or assume a bespoke configuration or non-standard process that a model could not reasonably have learned in general pre-training. \textit{Annotation indicator:} Annotators mark ``yes'' when the correct answer cannot be reliably inferred from general domain knowledge and requires awareness of document-specific context.}

\vspace{0.5em}
\noindent\textbf{Annotation Procedure.}

Two domain experts independently assessed each question against the three criteria above. For each criterion, the annotator assigned a binary label (``yes'' or ``no''). A question was classified as a CDFR item and excluded from the main evaluation set if any single criterion received a ``yes'' label from either annotator. Through this process, a total of 745 CDFR questions were identified and excluded from the main analysis set.

\vspace{0.5em}
\noindent\textbf{Illustrative Example.}

The following pair illustrates the distinction between a non-CDFR question and a CDFR question at the L1 memory level:

\begin{table}[!h]
\centering
\small
\begin{tabularx}{\linewidth}{p{0.18\linewidth}X X}
\toprule
\textbf{Attribute} & \textbf{Non-CDFR Example} & \textbf{CDFR Example} \\
\midrule
Question
  & What is the primary function of a transformer in a power transmission system?
  & According to the experiment described in the paper, what was the measured efficiency of the prototype transformer under 85\% load? \\
\addlinespace
Triggered Criterion
  & None
  & Criterion 1 (Specific Data Dependency) and Criterion 3 (Implicit Context Dependency) \\
\addlinespace
Classification
  & Non-CDFR (retained)
  & CDFR (excluded) \\
\addlinespace
Rationale
  & Answerable from general electrical engineering knowledge without reference to any specific document.
  & Requires access to a specific numerical measurement (efficiency value) from a particular prototype experiment described only in the source document. \\
\bottomrule
\end{tabularx}
\end{table}
\subsection{Validation Status Field Definitions}

Each sample in the IndustrialKnowEval dataset is annotated with a \texttt{validation\_status} field that records its passage status throughout the automated and manual verification pipeline. The field takes one of the following five values:

\begin{table}[h]
\centering
\small
\begin{tabularx}{\linewidth}{p{0.2\linewidth}p{0.25\linewidth}X}
\toprule
\textbf{Status Value} & \textbf{Stage} & \textbf{Definition} \\
\midrule
\texttt{auto\_pass}
  & Automated Verification
  & The sample successfully passed automated verification. \\
\addlinespace
\texttt{manual\_pass}
  & Manual Review
  & The logical chain of the sample and the evidentiary support were confirmed as reasonable during manual review. \\
\addlinespace
\texttt{auto\_fail} / \texttt{manual\_fail}
  & Automated Verification / Manual Review
  & The sample failed verification at either the automated stage (\texttt{auto\_fail}) or the manual review stage (\texttt{manual\_fail}).  \\
\addlinespace
\texttt{revised}
  & Post-Review Revision
  & The sample was initially assigned \texttt{auto\_fail} or \texttt{manual\_fail} status, subsequently underwent targeted revision by the research team or domain experts, and successfully passed re-verification. \\
\bottomrule
\end{tabularx}
\end{table}

\noindent\textit{Note: We discarded some samples labeled \texttt{auto\_fail} or \texttt{manual\_fail} without revising them, and these samples are not part of the final IndustrialKnowEval dataset. The released dataset keeps only samples whose final status is \texttt{auto\_pass}, \texttt{manual\_pass}, or \texttt{revised}. In the full quality control pipeline, automated verification removed 3,100 samples and manual review removed 2,222 samples. Another 7,266 samples passed re-validation after manual revision.}
\rev{\subsection{Filtering Statistics}}

\rev{Table~\ref{tab:filtering} reports the number of questions removed and retained at each filtering stage. One question per level was generated from each of the 12,151 source texts, and the final benchmark contains 10,945, 10,919, 10,930, and 10,488 questions at L1, L2, L3, and L4, respectively.}

\begin{table}[h]
\centering
\small
\caption{\rev{Questions removed and retained at each filtering stage.}}
\label{tab:filtering}
\resizebox{\linewidth}{!}{
\begin{tabular}{lcc}
\toprule
\textbf{Stage} & \textbf{Removed} & \textbf{Retained} \\
\midrule
Initial generation (12,151 per level) & -- & 48,604 \\
Automated verification & 3,100 & 45,504 \\
Manual review (including 745 CDFR exclusions) & 2,222 & 43,282 \\
\bottomrule
\end{tabular}
}
\end{table}

\section{\rev{Statistical Significance Tests and Macro-Averaged Results}}
\label{sec:stat_test}
\rev{
We compared GPT-5 with each of the other 17 models on the objective questions (L1--L3, $n=32{,}794$, with no exclusions) using paired bootstrap tests with 10,000 replicates and a fixed random seed (42). L4 open-ended questions were excluded because they are scored by an LLM judge. Because items derived from the same source document are not independent, the bootstrap resamples source documents (11,322 clusters). Knowledge fragments are nested within documents, and Table~\ref{tab:stat_test} also reports the results with fragment-level resampling for comparison. The tests are Bonferroni-corrected for the 17 comparisons ($\alpha=0.05/17=0.0029$). GPT-5 is significantly more accurate than 14 of the models and less accurate than GPT-6-sol, and its differences from Gemini~2.5~Pro and Qwen-Plus are not significant.
}

\begin{table}[!t]
\centering
\caption{Paired cluster bootstrap comparison of GPT-5 (accuracy 0.7458) with the other models on L1--L3 ($n=32{,}794$). $\Delta$Acc. is GPT-5 minus the model. The 95\% CI and $p_{\mathrm{doc}}$ use document-level resampling, and $p_{\mathrm{frag}}$ uses fragment-level resampling.}
\label{tab:stat_test}
\scriptsize\setlength{\tabcolsep}{1pt}
\begin{tabular*}{\linewidth}{@{\extracolsep{\fill}}lccccc@{}}
\toprule
\textbf{Model} & \textbf{Acc.} & \textbf{$\Delta$Acc.} & \textbf{95\% CI} & \textbf{$p_{\mathrm{doc}}$} & \textbf{$p_{\mathrm{frag}}$} \\
\midrule
GPT-6-sol & 0.8509 & $-$0.1051* & [$-$0.110, $-$0.101] & $<$0.001 & $<$0.001 \\
Gemini~2.5~Pro & 0.7428 & +0.0029 & [$-$0.002, 0.007] & 0.207 & 0.213 \\
Qwen-Plus & 0.7402 & +0.0056 & [0.001, 0.010] & 0.011 & 0.010 \\
GPT-4.1 & 0.7278 & +0.0180* & [0.014, 0.022] & $<$0.001 & $<$0.001 \\
DeepSeek-V3.1 & 0.7142 & +0.0315* & [0.027, 0.036] & $<$0.001 & $<$0.001 \\
DeepSeek-V4-Pro & 0.7121 & +0.0337* & [0.029, 0.038] & $<$0.001 & $<$0.001 \\
GLM-4.5 & 0.7057 & +0.0400* & [0.036, 0.044] & $<$0.001 & $<$0.001 \\
GLM-4.5-Air & 0.6991 & +0.0467* & [0.042, 0.051] & $<$0.001 & $<$0.001 \\
Qwen3-8B & 0.6740 & +0.0718* & [0.067, 0.076] & $<$0.001 & $<$0.001 \\
Qwen3-14B & 0.6657 & +0.0800* & [0.075, 0.085] & $<$0.001 & $<$0.001 \\
GLM-4-9B & 0.6557 & +0.0901* & [0.085, 0.095] & $<$0.001 & $<$0.001 \\
Qwen3-4B & 0.6514 & +0.0944* & [0.089, 0.099] & $<$0.001 & $<$0.001 \\
Llama-3.1-70B & 0.6503 & +0.0955* & [0.090, 0.101] & $<$0.001 & $<$0.001 \\
GLM-4 & 0.6365 & +0.1093* & [0.104, 0.114] & $<$0.001 & $<$0.001 \\
Qwen3-1.7B & 0.5304 & +0.2154* & [0.210, 0.221] & $<$0.001 & $<$0.001 \\
Llama-3.1-8B & 0.4793 & +0.2665* & [0.260, 0.273] & $<$0.001 & $<$0.001 \\
Qwen3-0.6B & 0.3545 & +0.3913* & [0.385, 0.398] & $<$0.001 & $<$0.001 \\
\bottomrule
\multicolumn{6}{l}{* Significant after Bonferroni correction ($\alpha=0.0029$).}
\end{tabular*}
\end{table}
\rev{
Tables~\ref{tab:macro_level} and~\ref{tab:macro_domain} report macro-averaged scores, in which each task type has equal weight, by cognitive level and by domain, respectively. L1--L3 are scored by accuracy and L4 by the normalized score of the primary judge. The rankings under micro and macro averaging are highly consistent (Spearman $\rho=0.998$).
}

\begin{table}[!t]
\centering
\caption{\rev{Macro-averaged scores by cognitive level. Overall is the mean over the 21 task types.}}
\label{tab:macro_level}
\scriptsize\setlength{\tabcolsep}{1pt}
\begin{tabular*}{\linewidth}{@{\extracolsep{\fill}}lccccc@{}}
\toprule
\textbf{Model} & \textbf{L1} & \textbf{L2} & \textbf{L3} & \textbf{L4} & \textbf{Overall} \\
\midrule
GPT-6-sol & 0.905 & 0.887 & 0.706 & 0.765 & 0.811 \\
GPT-5 & 0.785 & 0.801 & 0.585 & 0.847 & 0.746 \\
Gemini~2.5~Pro & 0.781 & 0.772 & 0.600 & 0.772 & 0.725 \\
Qwen-Plus & 0.772 & 0.797 & 0.581 & 0.739 & 0.715 \\
GPT-4.1 & 0.758 & 0.774 & 0.584 & 0.768 & 0.714 \\
DeepSeek-V3.1 & 0.753 & 0.767 & 0.551 & 0.773 & 0.703 \\
DeepSeek-V4-Pro & 0.770 & 0.783 & 0.498 & 0.795 & 0.701 \\
GLM-4.5 & 0.747 & 0.765 & 0.539 & 0.761 & 0.695 \\
GLM-4.5-Air & 0.741 & 0.755 & 0.518 & 0.754 & 0.684 \\
Qwen3-14B & 0.741 & 0.764 & 0.403 & 0.723 & 0.646 \\
Qwen3-8B & 0.709 & 0.736 & 0.478 & 0.685 & 0.644 \\
Qwen3-4B & 0.692 & 0.715 & 0.438 & 0.685 & 0.623 \\
GLM-4-9B & 0.685 & 0.698 & 0.488 & 0.620 & 0.617 \\
GLM-4 & 0.688 & 0.707 & 0.410 & 0.669 & 0.609 \\
Llama-3.1-70B & 0.696 & 0.717 & 0.448 & 0.560 & 0.598 \\
Qwen3-1.7B & 0.577 & 0.592 & 0.300 & 0.654 & 0.520 \\
Llama-3.1-8B & 0.481 & 0.499 & 0.383 & 0.435 & 0.446 \\
Qwen3-0.6B & 0.419 & 0.435 & 0.137 & 0.535 & 0.370 \\
\bottomrule
\end{tabular*}
\end{table}

\begin{table}[!t]
\centering
\caption{\rev{Macro-averaged scores by domain.}}
\label{tab:macro_domain}
\scriptsize\setlength{\tabcolsep}{1pt}
\begin{tabular*}{\linewidth}{@{\extracolsep{\fill}}lccccccc@{}}
\toprule
\textbf{Model} & \textbf{Aero.} & \textbf{Power} & \textbf{Metall.} & \textbf{Chemical} & \textbf{Auto.} & \textbf{Rail} & \textbf{Petrol.} \\
\midrule
GPT-6-sol & 0.841 & 0.832 & 0.773 & 0.806 & 0.822 & 0.811 & 0.793 \\
GPT-5 & 0.781 & 0.786 & 0.696 & 0.740 & 0.765 & 0.725 & 0.727 \\
Gemini~2.5~Pro & 0.743 & 0.763 & 0.683 & 0.713 & 0.735 & 0.715 & 0.710 \\
Qwen-Plus & 0.743 & 0.765 & 0.665 & 0.702 & 0.726 & 0.718 & 0.700 \\
GPT-4.1 & 0.735 & 0.757 & 0.665 & 0.714 & 0.733 & 0.682 & 0.687 \\
DeepSeek-V3.1 & 0.721 & 0.752 & 0.650 & 0.701 & 0.729 & 0.700 & 0.696 \\
DeepSeek-V4-Pro & 0.690 & 0.764 & 0.656 & 0.706 & 0.732 & 0.686 & 0.691 \\
GLM-4.5 & 0.710 & 0.739 & 0.650 & 0.688 & 0.721 & 0.684 & 0.682 \\
GLM-4.5-Air & 0.693 & 0.738 & 0.632 & 0.682 & 0.681 & 0.674 & 0.676 \\
Qwen3-14B & 0.642 & 0.701 & 0.597 & 0.650 & 0.658 & 0.658 & 0.633 \\
Qwen3-8B & 0.635 & 0.707 & 0.585 & 0.643 & 0.667 & 0.651 & 0.635 \\
Qwen3-4B & 0.607 & 0.709 & 0.577 & 0.614 & 0.643 & 0.615 & 0.612 \\
GLM-4-9B & 0.607 & 0.688 & 0.562 & 0.605 & 0.647 & 0.612 & 0.603 \\
GLM-4 & 0.605 & 0.667 & 0.551 & 0.605 & 0.622 & 0.610 & 0.598 \\
Llama-3.1-70B & 0.644 & 0.659 & 0.528 & 0.577 & 0.616 & 0.561 & 0.580 \\
Qwen3-1.7B & 0.503 & 0.611 & 0.455 & 0.513 & 0.545 & 0.530 & 0.517 \\
Llama-3.1-8B & 0.482 & 0.495 & 0.367 & 0.438 & 0.453 & 0.422 & 0.445 \\
Qwen3-0.6B & 0.345 & 0.448 & 0.320 & 0.370 & 0.389 & 0.352 & 0.363 \\
\bottomrule
\end{tabular*}
\end{table}

\clearpage
\section{Examples of Questions in IndustrialKnowEval}
\begin{example}{An Example of Aerospace L1 (CN)}
\begin{CJK}{UTF8}{gbsn}

\small
\textbf{difficulty:} L1

\textbf{question:}
星基增强系统(SBAS)主要目的是为了提升卫星导航系统的哪些性能?\\
\textbf{options:}
\\A) 精度,完好性,连续性和可用性 \\B) 速度,稳定性,覆盖性和兼容性 \\C) 灵敏度,抗干扰性,同步性和可维护性 \\D) 功耗,响应时间,扩展性和安全性
\\
\textbf{Correct Answer:} A\\
\textbf{Explanation:} 星基增强系统(SBAS)是在传统GNSS基础上发展而来，旨在满足民航等高安全要求用户对导航系统在精度、完好性、连续性和可用性方面的更高需求。选项B、C、D中的性能指标虽然在某些系统中重要，但不是SBAS设计所针对的核心性能提升目标。\\
\textbf{Knowledge\_reference:} "星基增强系统(SBAS)是在传统GNSS的基础上为了进一步满足民航用户对卫星导航系统越来越高的精度、完好性、连续性和可用性需求而应运而生。"\\
\textbf{Label:} 概念识记\\
\textbf{Validation\_status:} revised\\
\textbf{Category:} 航空航天\\
\end{CJK}
\end{example}

\begin{example}{An Example of Aerospace L1 (EN)}

\small
\textbf{difficulty:} L1
\textbf{question:}
What criterion is used to dynamically obtain and update the number of clusters in the described radar emitter classification framework?\\
\textbf{options:}
\\A) AIC criterion\\B) MDL criterion \\C) BIC criterion \\D) Cross-validation criterion
\\
\textbf{Correct Answer:} B\\
\textbf{Explanation:} The framework uses the MDL criterion to determine and update the number of clusters dynamically.\\
\textbf{Knowledge\_reference:} "The number of clusters can be obtained and updated dynamically by using MDL criterion rather than set-up in advance."\\
\textbf{Label:} Concept Recall\\
\textbf{Validation\_status:} auto\_pass\\
\textbf{Category:} Aerospace\\
\end{example}
\vfill\newpage
\begin{example}{An Example of Aerospace L2 (CN)}
\begin{CJK}{UTF8}{gbsn}

\small

\textbf{difficulty:} L2

\textbf{question:}
在弹载弹射与空中充气展升技术中，浮空器完成部署的主要步骤顺序是什么？\\
\textbf{options:}
\\A) 空中充气、弹射升空、减速伞展开、稳定驻空 \\B) 弹射升空、稳定驻空、空中充气、减速伞展开 \\C) 弹射升空、减速伞展开、空中充气、稳定驻空 \\D) 减速伞展开、弹射升空、稳定驻空、空中充气
\\
\textbf{Correct Answer:} C\\
\textbf{Explanation:} 正确的流程是首先通过弹射方式将浮空器送至高空，随后展开减速伞以控制姿态和速度，接着在空中进行充气展升，最终实现稳定驻空。选项A、B、D中的步骤顺序违背了物理实施逻辑，例如充气必须在减速和定位后进行。\\
\textbf{Knowledge\_reference:} " 建立了减速伞运动模型和伞一艇系统动力学模型, 通过仿真计算, 校验了方案可行性。"\\
\textbf{Label:} 工艺描述\\
\textbf{Validation\_status:} auto\_pass\\
\textbf{Category:} 航空航天\\
\end{CJK}
\end{example}

\begin{example}{An Example of Aerospace L2 (EN)}

\small
\textbf{difficulty:} L2
\textbf{question:}
What is the primary modeling challenge when developing an end-to-end model of a complex electro-mechanical-hydraulic system compared with single-domain models?\\
\textbf{options:}
\\A) Avoiding the use of component-level models\\B) Reducing the number of parameters to speed up simulation \\C) Ensuring only electrical components are represented \\D) Integrating multiple technical domains into a detailed, unified model 
\\
\textbf{Correct Answer:} D\\
\textbf{Explanation:} detailed modeling of the integrated end-to-end system presents significantly more challenge than conventional modeling limited to a single technical domain.\\
\textbf{Knowledge\_reference:} "Detailed modeling of the integrated end-to-end system presents significantly more challenge than the conventional modeling of components and system limited to single technical domain."\\
\textbf{Label:} Principle Explanation\\
\textbf{Validation\_status:} auto\_pass\\
\textbf{Category:} Aerospace\\
\end{example}
\vfill\newpage
\begin{example}{An Example of Aerospace L3 (CN)}
\begin{CJK}{UTF8}{gbsn}

\small
\textbf{difficulty:} L3

\textbf{question:}
相比天拓一号，天拓三号星载AIS系统在时间信息处理方面的改进带来了哪些关键优势？\\
\textbf{options:}
\\A) 实现了对每条报文的精确时间显示 \\B) 支持对海面船舶进行连续跟踪与监视 \\C) 显著降低了星上存储容量需求 \\D) 可生成AIS报文的发送时间用于航迹分析
\\
\textbf{Correct Answer:} A,B,D\\
\textbf{Explanation:} 天拓三号通过植入UTC时标实现了精确时间显示（A），为船舶连续跟踪与监视提供了技术基础（B），并可结合信号传播延迟生成发送时间，支持航迹优化分析（D）。C项错误，因为原文指出数据量增大会增加星上存储和带宽负荷，说明并未降低存储需求，反而带来挑战。\\
\textbf{Knowledge\_reference:} " 天拓三号AIS系统实现了对每一条报文的精确时间显示为今后实现对海面船舶进行连续跟踪与监视提供了良好的技术基础。由于AIS信号从发送到接收的延迟时间非常短...可生成每一条报文的发送时间,能够实现对船只的实时动态地连续跟踪和进行航迹优化分析等。然而这有可能带来数据冗余的问题,特别是当数据量特别大的时候,对星上存储容量、通信带宽等要求较高"\\
\textbf{Label:} 性能比较\\
\textbf{Validation\_status:} auto\_pass\\
\textbf{Category:} 航空航天\\
\end{CJK}
\end{example}

\begin{example}{An Example of Aerospace L3 (EN)}

\small
\textbf{difficulty:} L3

\textbf{question:}
What are plausible effects of dust-like solid carbon present during atmospheric entry into Titan on spacecraft systems?\\
\textbf{options:}
\\A) Improved radiative cooling of the TPS due to carbon's high emissivity\\B) High-temperature reactions with the TPS that degrade protection \\C) Formation of a cladding layer that enhances radiation absorption and overheats the TPS \\D) Blinding of onboard sensors by deposited carbon
\\
\textbf{Correct Answer:} B,C,D\\
\textbf{Explanation:} The text states that carbon can form a cladding layer favoring radiation absorption and overheating the TPS (C), can react with the TPS and spoil its protecting properties (B), and could blind sensors (D). It does not claim improved radiative cooling (A).\\
\textbf{Knowledge\_reference:} "(i) carbon can form acladding layer and might favorize radiation absorption, leading to an overheating of the TPS; (ii) high temperature solid carbon might react with the TPS and spoil its protecting properties; (iii) solid carbon could blind sensors."\\
\textbf{Label:} Influence Factor Analysis\\
\textbf{Validation\_status:} manual\_pass\\
\textbf{Category:} Aerospace\\
\end{example}
\vfill\newpage
\begin{example}{An Example of Aerospace L4 (CN)}
\begin{CJK}{UTF8}{gbsn}

\small
\textbf{difficulty:} L4

\textbf{question:}
如果某航天器流体管路系统出现压力异常下降，但无直接传感器监测泄漏，结合结构排除法，最可能的故障推理路径是什么？\\
\textbf{Correct Answer:} 首先利用系统结构关系分析可能的故障源，结合多个可测参数的变化情况排除无关分支，最终定位到导致压力下降的管路段，实现对泄漏位置的间接诊断。\\
\textbf{Knowledge\_reference:} " 对结构排除法进行了简单的介绍,在简单的例子上对利用结构排除法进行故障诊断做了示范,对应用多个测量参数进行诊断提出了简单易行的方法,提出了诊断管路泄漏的方法"\\
\textbf{Label:} 故障排查\\
\textbf{Validation\_status:} revised\\
\textbf{Category:} 航空航天\\
\end{CJK}
\end{example}

\begin{example}{An Example of Aerospace L4 (EN)}

\small
\textbf{difficulty:} L4

\textbf{question:}
Why is electron tube technology still considered appropriate for certain high-power applications, and what system-level reasoning supports maintaining tube-based solutions alongside low-power solid-state technologies?\\
\textbf{Correct Answer:} Electron tube technology remains appropriate because high-power "appropriate" technology historically preceded lower-power solid-state technology, and only after widespread adoption of low-power solid-state solutions have their limitations in high-power regimes become evident. Maintaining tube-based solutions is justified at the system level since modern microwave power sources and amplifiers-such as microwave power modules integrating solid-state drivers with miniature TWTs, multi stage-depressed-collector TWTs for satellite and EW use, Clustered-Cavity klystrons for radar, and MBKs with convergent electron guns-deliver bandwidth, efficiency, low-voltage operation, and long life needed for demanding radar and communications roles. Additionally, devices like the Constant Efficiency Amplifier can efficiently generate high-average-power pseudo-Gaussian pulses valuable for detecting stealth targets at long range, a capability challenging for purely low-power solid-state approaches.\\
\textbf{Knowledge\_reference:} "Tube technology is alive, vital, and is still appropriate technology for many applications. In my earlier analogies with engines, the lower-power internal combustion engine came first and was followed by the more powerful jet engine. The jet engine received the attention it deserved,partly because it was new."\\
\textbf{Label:} System Optimization\\
\textbf{Validation\_status:} auto\_pass\\
\textbf{Category:} Aerospace\\
\end{example}
\clearpage

\section{\rev{Error Analysis and Case Studies}}
\label{sec:case_study}
\rev{This appendix reports the error patterns of the 18 models across model families, cognitive levels, and task types, followed by representative cases from several domains and levels.}

\rev{\textbf{Error patterns across families.} Table~\ref{tab:family_errors} classifies each incorrect L3 answer by comparing it with the reference set: under-selection omits correct options, over-selection adds incorrect ones, and other errors do both. The families differ in how they err at L3. Over-selection accounts for 54--70\% of the L3 errors of the GPT, GLM, and Qwen families, but under-selection accounts for 51\% of those of the DeepSeek family, and Gemini and Llama fall in between. The difficulty profile across task types is shared by all models: data extraction is the most error-prone L1 task type for each of the 18 models (error rate 0.26 for GPT-6-sol and 0.43 for GPT-5, compared with 0.04--0.07 and 0.10--0.22 for their other L1 task types), and formula comprehension is the most error-prone L2 task type for each model.}

\rev{\textbf{Task difficulty.} Of the L3 items, 8.2\% are solved by none of the 18 models and 3.2\% by all of them, compared with 0.1\% and 20.1\% of the L1 items. For 6.8\% of the L1--L3 items, at least 12 models choose the same incorrect answer. The share of source documents whose questions are answered with a mean accuracy below 0.5 is 31.7\% in Metallurgy, compared with 8.5\% in Power.}

\rev{\textbf{Why Metallurgy is more difficult.} The task composition of Metallurgy differs markedly from that of the other domains: data extraction accounts for 751 of its 1,731 L1 items (43\%), compared with 137 of 1,754 (8\%) in Power. After each domain is reweighted to a common task-type distribution within each level, the L1--L3 accuracy gap between Metallurgy and the mean of the other domains decreases from 9.1 to 5.4 percentage points on average over the 18 models, so task composition accounts for about 40\% of the gap (30--52\% for individual models). Within data extraction, the accuracy of Metallurgy is comparable to that of the other domains (0.485 vs.\ 0.487). The remaining gap is concentrated in conceptual and process-related task types, such as influence factor analysis at L3 (0.386 vs.\ 0.489) and process description at L2 (0.615 vs.\ 0.713).}

\begin{table}[!t]
\centering
\caption{\rev{Accuracy by level and L3 error pattern by model family, pooled over the member models. Under, Over, and Other are shares of the L3 errors.}}
\label{tab:family_errors}
\scriptsize\setlength{\tabcolsep}{1pt}
\begin{tabular*}{\linewidth}{@{\extracolsep{\fill}}lccccccc@{}}
\toprule
\textbf{Family} & \textbf{Models} & \textbf{L1} & \textbf{L2} & \textbf{L3} & \textbf{Under} & \textbf{Over} & \textbf{Other} \\
\midrule
GPT & 3 & 0.810 & 0.871 & 0.644 & 15.3\% & 69.5\% & 15.2\% \\
Gemini & 1 & 0.773 & 0.836 & 0.619 & 33.0\% & 44.6\% & 22.4\% \\
DeepSeek & 2 & 0.755 & 0.834 & 0.550 & 51.1\% & 28.6\% & 20.3\% \\
GLM & 4 & 0.713 & 0.791 & 0.519 & 15.6\% & 61.8\% & 22.6\% \\
Llama & 2 & 0.591 & 0.661 & 0.443 & 32.5\% & 37.1\% & 29.8\% \\
Qwen & 6 & 0.654 & 0.735 & 0.419 & 16.0\% & 54.0\% & 29.9\% \\
\bottomrule
\end{tabular*}
\end{table}

\textbf{Case studies.} Power is the best-performing domain, with a mean L1--L3 accuracy of 76.5\% and a mean L4 score of 7.38/10 over the 18 models.

\begin{case}{Case 1: High-Score Example (Score: 10/10)}
\textbf{Question :} Why is energy storage optimization critical for power system reliability under high renewable penetration?\\

\textbf{Reference Answer:} Because renewable energy generation is intermittent and volatile, easily causing load imbalance and supply instability. Energy storage can smooth fluctuations, respond to load changes, and ensure continuous supply to enhance system resilience.

\textbf{GPT-5 Answer:} [Comprehensive response covering: dispatchability compensation for stochastic wind/solar output; peak shaving and valley filling at minute-to-hour timescales; reserve and black-start capabilities; frequency and voltage support through primary/secondary regulation...]

\textbf{Analysis:} GPT-5 provided a complete and technically accurate response that fully addressed the question while extending beyond the reference answer with additional valid points about grid stability mechanisms.
\end{case}

Metallurgy is the most challenging domain, with a mean L1--L3 accuracy of 58.3\% and a mean L4 score of 6.34/10.

\begin{case}{Case 2: Low-Score Example (Score: 2/10)}
\begin{CJK}{UTF8}{gbsn}
\textbf{Question (CN):} 如果在红外光谱检测中发现νc=0峰强度明显增强，最可能的根本原因是什么？\\
\textit{(If a significant increase in the intensity of the νc=0 peak is observed during infrared spectroscopy detection, what is the most likely root cause?)}

\textbf{Reference Answer:} The intensification of the νc=0 peak indicates an increase in the content of carbonyl (C=O) functional groups. This may result from a higher relative proportion of ester components (such as sebacate-based base oils) in the base oil or the occurrence of oxidation reactions generating carbonyl-containing degradation products.

\textbf{GPT-5 Answer:} The most probable root cause is an artificial enhancement of the signal around the νc=0 region, resulting from a heightened baseline due to sources of radiation or scattering near zero frequency (ν≈0). This can arise from strong thermal emission from the sample, stray or background light leakage within the instrument, detector saturation leading to baseline drift, or increased scattering from optical components or the sample itself concentrated at the low-frequency end. Ultimately, the issue is attributed to an amplified background signal of non-absorption origin rather than a genuine intrinsic spectral band.

\textbf{Analysis:} The incorrect analysis primarily attributes the intensified νc=0 peak to artificial signal amplification caused by a raised baseline from non-absorption sources like thermal emission, stray light, or detector saturation near zero frequency. It treats the spectral change as an instrumental artifact rather than relating peak intensity to the concentration of the functional group.
\end{CJK}
\end{case}

\begin{case}{Case 3: Process Mechanism Comprehension}
\begin{CJK}{UTF8}{gbsn}
\textbf{Question (CN):} 在Pr$_{60}$Tb$_{20}$Al$_{10}$Ga$_{10}$晶界扩散过程中，Pr元素的作用主要体现在哪个方面？\\
\textit{(In grain boundary diffusion, what is the primary role of Pr element?)}

\textbf{Options:} A. 提高主相晶粒尺寸; B. 改善晶界相润湿性; C. 增加磁体密度; D. 替代Nd元素分布
\textit{(A. Increase the grain size of the main phase；B. Improve the wettability of the grain boundary phase；C. Increase the magnet density；D. Substitute for Nd in the microstructural distribution)}

\textbf{Ground Truth:} B \quad \textbf{GPT-5 Answer:} D \quad \textbf{Error Rate:} 4/18 models

\textbf{Analysis:} In the Pr$_{60}$Tb$_{20}$Al$_{10}$Ga$_{10}$grain‐boundary diffusion process, the dominant function of Pr is to depress the eutectic temperature and thereby enhance the liquid-phase wettability along grain boundaries, which facilitates more efficient infiltration and intergranular phase redistribution. GPT-5 instead selected substitution for Nd, a common role of rare-earth elements in Nd-Fe-B magnets, rather than the mechanism specific to this alloying system.
\end{CJK}
\end{case}
\begin{case}{Case 4: Safety Assessment Error (Universal Failure)}
\textbf{Question:} According to TOPSIS analysis, which factors are identified as key threats to underground mine safety?\\
\textbf{Options:} A. Operational status of the six major systems; B. Geological conditions of the mine; C. Workplace safety management; D. Employee health conditions

\textbf{Ground Truth:} A, B, C \quad \textbf{GPT-5 Answer:} A, B \quad \textbf{Error Rate:} 14/18 models 

\textbf{Analysis:} The models select different incomplete or excessive subsets: GPT-5 omits workplace safety management (C), DeepSeek-V3.1 omits the operational status of the six major systems (A), and Qwen-Plus also adds employee health conditions (D). Identifying the exact set requires the specific result of this assessment, which is difficult to infer from general knowledge of mine safety.
\end{case}

\begin{case}{Case 5: Incomplete Reasoning}
\textbf{Question:} Why is it problematic to restrict new designs to compatibility with existing systems when developing future space exploration vehicles?

\textbf{Options:} A. It prevents taking advantage of new hardware and software technologies; B. It increases life cycle costs due to custom compiler and runtime needs when extending ATLAS; C. It reduces flexibility compared to commercial ADEs that support multi-threaded programming; D. It eliminates the structured environment required for long-term maintenance

\textbf{Ground Truth:} A, B, C 

\textbf{GPT-5 Answer:} A, C \quad \textbf{Error Rate:} 7/18 models

\textbf{Analysis:} GPT-5 identifies the loss of access to new technologies (A) and of flexibility (C) but omits the life-cycle cost of custom compilers and runtime support (B), which illustrates the under-selection that exact set matching counts as incorrect.

\end{case}

\begin{case}{Case 6: Factual Recall}
\begin{CJK}{UTF8}{gbsn}
\textbf{Question (CN):} 制定动力电池领域元数据标准时可参照的原则来自哪个标准文件？\\
\textit{(Which standard document provides the guiding principles for formulating metadata standards in the power battery domain?)}

\textbf{Options:} A. SDS/T 2111-2004; B. SDS/T 2112-2005; C. SDS/T 2113-2006; D. SDS/T 2114-2007

\textbf{Ground Truth:} A \quad \textbf{GPT-5 Answer:} B \quad \textbf{Error Rate:} 13/18 models

\textbf{Analysis:} This item requires precise recollection of China-specific metadata principles and standards for power batteries (SDS/T 2111-2004). However, GPT-5 selected SDS/T 2112-2005 instead, reflecting a gap in its training data coverage. Since Chinese national or industry standards are relatively limited in number, this led to an incorrect selection.
\end{CJK}
\end{case}

In 1,276 of the L1--L3 items, GPT-5 errs but both DeepSeek-V3.1 and Qwen-Plus are correct; Case~7 is an example.

\begin{case}{Case 7: Numerical Parameter Error}
\begin{CJK}{UTF8}{gbsn}

\textbf{Question (CN):} 在利用赤泥制备的PAC基液处理含锰酸性废水时，适宜的pH值是多少？\\
\textit{(What is the optimal pH for treating manganese-containing acidic wastewater with red mud-based PAC?)}

\textbf{Options:} A. pH=4; B. pH=7; C. pH=8; D. pH=10

\textbf{Ground Truth:} C (pH=8) \quad \textbf{GPT-5:} A (pH=4)  \quad \textbf{DeepSeek-V3.1:} C (pH=8) \checkmark \quad \textbf{Qwen-Plus:} C (pH=8) \checkmark

\textbf{Analysis:} GPT-5’s incorrect selection of pH=4 reflects a parameter-level deviation rather than a mechanism-level misunderstanding: the model activated general knowledge about acidic wastewater characteristics instead of retrieving the specific operational window associated with PAC coagulation chemistry. By contrast, DeepSeek and Qwen-Plus correctly retrieved the canonical pH value, demonstrating more stable recall of metallurgy-related treatment parameters.
\end{CJK}
\end{case}
\clearpage

\section{Controlled Subsets}
\label{sec:controlled_subsets}

This appendix gives the construction rules, the grading rules, and the prompts of the two controlled subsets used in Section~V-C. The results are reported in Section~V-C. The subsets match question format across levels in opposite directions: all items are single-answer questions in one subset and open-ended questions in the other. Each subset contains 50 items per level from each of the seven domains, with the 50 Aerospace items divided equally between the Chinese and English versions, that is, 350 items per level and 1,400 items in total. The source items were randomly sampled from the main benchmark within each domain and level, and the two subsets were sampled independently, so that they share no source items.

\subsection{Single-Choice Subset}

L1 and L2 items are taken from the main benchmark without modification, and L3 and L4 items were converted into single-answer four-option questions (700 converted items).

\textbf{L3 conversion.} The four candidate statements of the original item are listed in the stem. The correct option is the exact set of correct statements, and each distractor is obtained from this set by omitting one correct statement, adding one incorrect statement, or replacing one correct statement with an incorrect one. This conversion is rule-based and introduces no new content.

\textbf{L4 conversion.} The original question is retained verbatim, and the correct option condenses the reference answer. Three distractors were written by Claude (claude-opus-5-5) as plausible but incorrect answers, for example by invoking a related mechanism that does not apply, by giving an incomplete or reversed causal chain, or by contradicting a specific value or object in the reference answer.

\textbf{Automatic checks.} In the converted items, the position of the correct option is balanced across A to D. The options were checked so that the correct option cannot be identified from its length, from the presence of numbers, or from wording that refers to the source document. All items are scored by accuracy.

\subsection{Open-Ended Subset}

L1, L2, and L3 items were converted into open-ended questions by Qwen-Plus, with reference answers derived from the correct options of the original items (1,050 converted items), using the prompt below. L4 items are already open-ended and are used without modification. Each converted L3 question states the number of points required, which equals the number of correct statements of the original item.

\begin{prompt}{Prompt for Converting L1--L3 Items into Open-Ended Questions}
You are an expert item writer in the industrial domain. Rewrite the following \{kind\} as an open-ended question and give its reference answer.
\\
\\
{[}Original level{]}: \{level\} (\{level\_def\})

{[}Original task type{]}: \{label\}

{[}Original question{]}: \{question\}

{[}Options{]}: \{options\}

{[}Correct option(s){]}: \{answer\}

{[}Source passage{]}: \{reference\}
\\
\\
Requirements:

1. Test the same knowledge point and require the same reasoning as the original question, so that the cognitive level is unchanged. \{level\_rule\}

2. Do not list options in the question, and do not reveal the answer or hint at its range in the question.

3. The question must have a definite, checkable answer; the reference answer must be directly supported by the [Correct option(s)] and the [Source passage] and must not add information absent from the passage.

4. The question and reference answer must be self-contained, with no references such as "this paper", "the study", "the above", or "the options".
\\
\\
Output only one JSON object and nothing else:

\{"question": "...", "answer": "..."\}
\end{prompt}

The level rule states that the reference answer is a short fact, value, or term at L1, one or two sentences explaining the concept, principle, or process at L2, and the points of all correct options at L3.

All conversions kept the knowledge point and the reasoning required by the original item. Three experts reviewed every converted item and confirmed that its level had not changed under the definitions in Section~III. Experts from each domain also took part in this review. An item entered a subset only after the experts approved it.

\subsection{Grading of the Open-Ended Subset}

\textbf{L4 rubric.} All items are graded with the prompt and rubric used for L4 in the main benchmark (Appendix~C). The judge is Qwen-Plus, with the same decoding settings as the primary judge. GPT-6-sol serves as a second judge.

\textbf{All-or-nothing rule.} The L4 rubric awards graded scores from 1 to 10, so it gives partial credit at every level, but L3 in the main benchmark gives none. We therefore also graded the subset with a binary rule applied identically to all levels, which is the open-ended counterpart of exact set match. Before it was applied to the full subset, the rule was tested on a pilot sample of 160 items (five per domain and level), and three experts reviewed the verdicts and confirmed the rule, with experts from each domain also taking part. The two rules measure different things: of the 195 L4 responses of Qwen-Plus that score at least 8/10 under the L4 rubric, 132 are graded incorrect under the all-or-nothing rule.

\begin{prompt}{Prompt for All-or-Nothing Grading}
You are an expert answer grader in the industrial domain. Based on the [Question] and the [Reference Answer], decide whether the [Model Answer] is correct.
\\
\\
Grading rules (applied identically to all questions):

1. First identify all key points of the [Reference Answer] (facts, values, objects, causal steps).

2. Grade "Correct" only if the [Model Answer] covers every key point and contains no factual or reasoning error that contradicts the [Reference Answer]; otherwise grade "Incorrect".

3. Wording equivalent in meaning to the [Reference Answer] counts as covering that key point; literal match is not required. Additional content in the [Model Answer] that does not contradict the [Reference Answer] does not affect the grade.

4. Covering only some key points, giving a different explanation or mechanism from the [Reference Answer], or any step of the reasoning chain contradicting the [Reference Answer] is graded "Incorrect".

5. Ignore answer length and style.
\\
\\
{[}Question{]}: \{question\}

{[}Reference Answer{]}: \{correct\_answer\}

{[}Model Answer{]}: \{model\_answer\}
\\
\\
Output in this format:

Verdict: Correct or Incorrect

Reason: one sentence naming the missing or wrong key point; if correct, name the key points covered
\end{prompt}

\section{\rev{ICCG Variants}}
\label{sec:iccg_variants}

\rev{This appendix compares the revised ICCG with alternative gap metrics. Let $A_{12}=\mathrm{Acc}(\mathrm{L1{+}L2})$ and $H=\mathrm{Acc}(\mathrm{L3{+}L4})$. The revised metric is $\mathrm{ICCG}=1-H/A_{12}$. The variants are the absolute difference $\Delta=A_{12}-H$, the L3-only variant $\mathrm{ICCG}_{\mathrm{L3}}=1-\mathrm{Acc}(\mathrm{L3})/A_{12}$, which uses only objectively scored items, and ICCG with L4 scored by GPT-6-sol or Qwen3.7-Plus. Each value is computed per domain and averaged over the seven domains.}

\rev{Table~\ref{tab:iccg_variants} lists the values for each model, and Table~\ref{tab:iccg_rank} compares the model rankings. The exponent used in the original form, $1-r^{\gamma}$, is strictly decreasing in $r$ for any $\gamma>0$, so the ranking within each domain is the same for every $\gamma$, and for the mean across domains the Spearman correlation with the revised ICCG rounds to 1.00 for $\gamma\in\{2,3,4,5\}$. The gap quantities capture complementary aspects of a profile, and ICCG is therefore reported together with $A_{12}$ and $H$.}

\begin{table}[!t]
\centering
\caption{\rev{ICCG and its variants for each model, ordered by ICCG. $H$ is $\mathrm{Acc}(\mathrm{L3{+}L4})$. G6 and Q3.7 denote ICCG with L4 scored by GPT-6-sol and Qwen3.7-Plus.}}
\label{tab:iccg_variants}
\scriptsize\setlength{\tabcolsep}{1pt}
\begin{tabular*}{\linewidth}{@{\extracolsep{\fill}}lccccccc@{}}
\toprule
\textbf{Model} & $A_{12}$ & $H$ & \textbf{ICCG} & $\Delta$ & $\mathrm{ICCG}_{\mathrm{L3}}$ & \textbf{G6} & \textbf{Q3.7} \\
\midrule
GPT-5 & 0.815 & 0.722 & 0.113 & 0.093 & 0.262 & 0.238 & 0.200 \\
GPT-4.1 & 0.788 & 0.680 & 0.137 & 0.108 & 0.236 & 0.234 & 0.219 \\
Gemini~2.5~Pro & 0.804 & 0.693 & 0.137 & 0.110 & 0.229 & 0.257 & 0.241 \\
DeepSeek-V3.1 & 0.786 & 0.669 & 0.148 & 0.117 & 0.271 & 0.260 & 0.253 \\
GLM-4.5 & 0.777 & 0.658 & 0.153 & 0.119 & 0.280 & 0.262 & 0.249 \\
GLM-4.5-Air & 0.775 & 0.648 & 0.164 & 0.127 & 0.297 & 0.283 & 0.269 \\
Qwen-Plus & 0.807 & 0.667 & 0.174 & 0.140 & 0.251 & 0.266 & 0.257 \\
DeepSeek-V4-Pro & 0.803 & 0.656 & 0.184 & 0.147 & 0.339 & 0.254 & 0.255 \\
GPT-6-sol & 0.913 & 0.744 & 0.186 & 0.170 & 0.208 & 0.175 & 0.212 \\
Llama-3.1-8B & 0.510 & 0.411 & 0.196 & 0.100 & 0.196 & 0.288 & 0.327 \\
Qwen3-8B & 0.752 & 0.597 & 0.207 & 0.155 & 0.311 & 0.294 & 0.292 \\
Qwen3-1.7B & 0.626 & 0.493 & 0.213 & 0.133 & 0.458 & 0.343 & 0.352 \\
Qwen3-4B & 0.736 & 0.580 & 0.213 & 0.156 & 0.345 & 0.316 & 0.317 \\
GLM-4-9B & 0.721 & 0.567 & 0.214 & 0.154 & 0.272 & 0.306 & 0.323 \\
GLM-4 & 0.732 & 0.550 & 0.251 & 0.182 & 0.402 & 0.348 & 0.359 \\
Qwen3-14B & 0.777 & 0.578 & 0.257 & 0.199 & 0.433 & 0.352 & 0.346 \\
Qwen3-0.6B & 0.464 & 0.329 & 0.280 & 0.135 & 0.705 & 0.436 & 0.442 \\
Llama-3.1-70B & 0.739 & 0.504 & 0.319 & 0.235 & 0.362 & 0.390 & 0.405 \\
\midrule
Mean & 0.740 & 0.597 & 0.197 & 0.143 & 0.325 & 0.295 & 0.295 \\
\bottomrule
\end{tabular*}
\end{table}

\begin{table}[!t]
\centering
\caption{\rev{Robustness of model rankings to the gap metric, based on the mean across domains. $\rho$ is the Spearman correlation with the revised ICCG; Smallest and Largest name the models with the smallest and largest gaps.}}
\label{tab:iccg_rank}
\scriptsize\setlength{\tabcolsep}{1pt}
\begin{tabular*}{\linewidth}{@{\extracolsep{\fill}}lccc@{}}
\toprule
\textbf{Gap Metric} & \textbf{$\rho$} & \textbf{Smallest} & \textbf{Largest} \\
\midrule
Revised ICCG, $1-r$ & 1.00 & GPT-5 & Llama-3.1-70B \\
Absolute difference $\Delta$ & 0.779 & GPT-5 & Llama-3.1-70B \\
$\mathrm{ICCG}_{\mathrm{L3}}$ (without L4) & 0.692 & Llama-3.1-8B & Qwen3-0.6B \\
ICCG, L4 scored by GPT-6-sol & 0.880 & GPT-6-sol & Qwen3-0.6B \\
ICCG, L4 scored by Qwen3.7-Plus & 0.897 & GPT-5 & Qwen3-0.6B \\
\bottomrule
\end{tabular*}
\end{table}

\vfill\newpage
\section{\rev{Comparison Across Generation Models}}
\label{app:generator}

\rev{All items of the main benchmark were generated by Qwen-Plus. Because Qwen models are also among the evaluated models, this appendix tests whether the choice of generator affects the model ranking and whether models of the generator's family are favored.}

\rev{\subsection{Item Generation}}

\rev{We randomly sampled 100 source texts from the Chinese sources of the seven domains (15 each from Power and Metallurgy and 14 each from the other five domains; random seed 42). L4 was not included in this comparison.}

\rev{For each text, Claude Opus 5.5 (claude-opus-5-5) generated one L1, one L2, and one L3 item with the same input and template as the original generation, which yielded 300 items. Explanations and knowledge references were not generated. Among the new L3 items, 44 have two correct statements and 56 have three, compared with 45, 54, and 1 (four correct statements) among the 100 original L3 items.}

\rev{\subsection{Quality Control}}

\rev{The new items first passed an automatic check of forbidden context-dependent expressions, four distinct options, option length, and, for L3, at least two correct statements. One item was flagged and corrected. The items were then reviewed by three experts (one from industry and two working in industrial modeling and simulation), who raised no objections, so all 300 items were retained without further change.}

\rev{\subsection{Evaluation}}

\rev{We evaluated 12 of the 18 models on the new items: GPT-6-sol, Qwen-Plus, Gemini~2.5~Pro, GPT-4.1, DeepSeek-V4-Pro, DeepSeek-V3.1, GPT-5, GLM-4.5, GLM-4.5-Air, Qwen3-14B, Llama-3.1-70B, and Llama-3.1-8B. Qwen3-0.6B, Qwen3-1.7B, Qwen3-4B, and Qwen3-8B could no longer be accessed through the provider, and GLM-4 and GLM-4-9B were not included. The new items used the zero-shot prompts of the main benchmark, the options were shuffled once per item, and L3 was scored by exact-set matching.}

\rev{Qwen-Plus was evaluated with qwen-plus-2025-12-01 on both versions because the version used in the main benchmark (qwen-plus-2025-09-11) could no longer be accessed. Confidence intervals are percentile bootstrap intervals with source texts as resampling units.}

\rev{\subsection{Results}}

\begin{table}[!t]
\centering
\caption{\rev{Accuracy on the same 100 source texts (300 L1--L3 items) with Qwen-Plus and Claude Opus 5.5 as generators. The difference is Qwen minus Claude, with a 95\% bootstrap confidence interval.}}
\label{tab:generator-models}
\scriptsize\setlength{\tabcolsep}{1pt}
\begin{tabular*}{\linewidth}{@{\extracolsep{\fill}}lccc@{}}
\toprule
Model & Qwen & Claude & Difference [95\% CI] \\
\midrule
GPT-6-sol & 0.890 & 0.843 & +0.047 [$-$0.010, +0.103] \\
Qwen-Plus & 0.787 & 0.800 & $-$0.013 [$-$0.080, +0.057] \\
Gemini~2.5~Pro & 0.777 & 0.783 & $-$0.007 [$-$0.067, +0.053] \\
GPT-4.1 & 0.773 & 0.773 & 0.000 [$-$0.067, +0.067] \\
DeepSeek-V4-Pro & 0.760 & 0.783 & $-$0.023 [$-$0.090, +0.043] \\
DeepSeek-V3.1 & 0.760 & 0.717 & +0.043 [$-$0.027, +0.113] \\
GPT-5 & 0.760 & 0.790 & $-$0.030 [$-$0.100, +0.040] \\
GLM-4.5 & 0.750 & 0.790 & $-$0.040 [$-$0.107, +0.030] \\
GLM-4.5-Air & 0.730 & 0.770 & $-$0.040 [$-$0.113, +0.030] \\
Qwen3-14B & 0.720 & 0.750 & $-$0.030 [$-$0.100, +0.040] \\
Llama-3.1-70B & 0.717 & 0.663 & +0.053 [$-$0.023, +0.130] \\
Llama-3.1-8B & 0.520 & 0.477 & +0.043 [$-$0.027, +0.113] \\
\bottomrule
\end{tabular*}
\end{table}

\rev{Table~\ref{tab:generator-models} lists the results by model. The model rankings under the two generators are consistent, with a Spearman correlation of $\rho=0.79$ (95\% CI [0.37, 0.91]; L1--L2: 0.62 [0.21, 0.86]; L3: 0.64 [0.30, 0.87]). GPT-6-sol ranks highest and Llama-3.1-8B lowest under both generators. The interval is wide, and individual positions in the middle of the ranking change.}

\rev{The two Qwen models score 0.022 lower on Qwen-generated items than on Claude-generated items, but the other ten models score 0.005 higher. The difference between these two quantities is $-0.026$ (95\% CI [$-$0.064, 0.011]), so we detect no preference of Qwen models for items generated by Qwen-Plus.}

\end{document}